\documentclass{article} 
\usepackage{iclr2022_conference,times}

\usepackage{amsmath,amsfonts,bm}

\def\eqref#1{equation~\ref{#1}}

\def\1{\bm{1}}

\def\rk{{\textnormal{k}}}

\def\rq{{\textnormal{q}}}

\def\rvr{{\mathbf{r}}}

\def\rmR{{\mathbf{R}}}

\def\va{{\bm{a}}}
\def\vb{{\bm{b}}}
\def\vc{{\bm{c}}}

\def\vk{{\bm{k}}}

\def\vm{{\bm{m}}}

\def\vp{{\bm{p}}}
\def\vq{{\bm{q}}}

\def\vv{{\bm{v}}}

\def\vx{{\bm{x}}}
\def\vy{{\bm{y}}}

\def\mA{{\bm{A}}}
\def\mB{{\bm{B}}}
\def\mC{{\bm{C}}}

\def\mI{{\bm{I}}}

\def\mK{{\bm{K}}}

\def\mM{{\bm{M}}}

\def\mQ{{\bm{Q}}}
\def\mR{{\bm{R}}}

\def\mU{{\bm{U}}}
\def\mV{{\bm{V}}}
\def\mW{{\bm{W}}}
\def\mX{{\bm{X}}}
\def\mY{{\bm{Y}}}

\DeclareMathAlphabet{\mathsfit}{\encodingdefault}{\sfdefault}{m}{sl}
\SetMathAlphabet{\mathsfit}{bold}{\encodingdefault}{\sfdefault}{bx}{n}
\newcommand{\tens}[1]{\bm{\mathsfit{#1}}}

\def\tW{{\tens{W}}}
\def\tX{{\tens{X}}}
\def\tY{{\tens{Y}}}

\newcommand{\softmax}{\mathrm{softmax}}

\usepackage{hyperref}
\usepackage{url}
\usepackage{booktabs}
\usepackage{thmtools} 
\usepackage{thm-restate}
\usepackage{graphicx,floatrow}
\usepackage{array,multirow}

\usepackage{graphicx}
\usepackage{caption}
\usepackage{subcaption}
\graphicspath{ {./figures/} }
\usepackage{thmtools} 
\usepackage{thm-restate}
\usepackage{tikz}

\newcommand{\uuline}[1]{%
    \tikz[baseline=(todotted.base)]{
        \node[inner sep=1pt,outer sep=0pt] (todotted) {#1};
        \draw (todotted.south west) -- (todotted.south east);
        \draw ([yshift=-1.2pt]todotted.south west) -- ([yshift=-1.2pt]todotted.south east);
    }%
}%

\title{FastRPB: a Scalable Relative Positional Encoding for Long Sequence Tasks}

\author{Maksim Zubkov \\
Moscow Institute of Physics and Technology \\
VK Lab \\
\texttt{zubkov.md@phystech.edu} \\
\And
Daniil Gavrilov \\
Tinkoff \\
\texttt{d.gavrilov@tinkoff.ru}
}

\iclrfinalcopy 
\begin{document}

\maketitle

\begin{abstract}
Transformers achieve remarkable performance in various domains, including NLP, CV, audio processing, and graph analysis. However, they do not scale well on long sequence tasks due to their quadratic complexity w.r.t. the input’s length. Linear Transformers were proposed to address this limitation. However, these models have shown weaker performance on the long sequence tasks comparing to the original one. 

In this paper, we explore Linear Transformer models, rethinking their two core components. Firstly, we improved Linear Transformer with  \textbf{S}hift-\textbf{I}nvariant \textbf{K}ernel \textbf{F}unction \textbf{SIKF}, which achieve higher accuracy without loss in speed. Secondly, we introduce \textbf{FastRPB}\footnote{The work on this paper was started at the end of 2020 and finished in fall 2021. However, right before us \citet{one} was published, which uses a similar scheme to utilize relative positional information with FFT. Although we still believe that one could derive insights from our work, we publish it as a preprint.} which stands for \textbf{Fast} \textbf{R}elative \textbf{P}ositional \textbf{B}ias, which efficiently adds positional information to self-attention using Fast Fourier Transformation. FastRPB is independent of the self-attention mechanism and can be combined with an original self-attention and all its efficient variants. FastRPB has $\mathcal{O}(N\log{N})$ computational complexity, requiring $\mathcal{O}(N)$ memory w.r.t. input sequence length $N$. 

We compared introduced modifications with recent Linear Transformers in different settings: text classification, document retrieval, and image classification. Extensive experiments with FastRPB and SIKF demonstrate that our model significantly outperforms another efficient positional encodings method in accuracy, having up to x1.5 times higher speed and requiring up to x10 times less memory than the original Transformer. 
\end{abstract}

\section{Introduction}

Transformer architecture~\citep{vaswani2017attention} originally proposed for machine translation tasks has shown impressive results in a wide range of domains, including natural language processing, image recognition, audio captioning, graph analysis, and bioinformatics~\citep{lin2021survey}. However, in applications that require processing long sequences, the benefits of transformers are often accompanied by high consumption of computational and memory resources. The main bottleneck is the transformer's core component, the self-attention mechanism. Self-attention computes similarity scores for all pairs of tokens in the input sequence, and therefore, it has a quadratic complexity $\mathcal{O}(N^2)$ in computations and memory relative to the length of the input sequence $N$\footnote{The full complexity of self-attention also depends on attention head size $D$. For the original self-attention, complexity is $\mathcal{O}(N^2D)$}.

Recently, several approaches have been introduced to reduce the computational complexity and memory footprint of self-attention. Some works utilize the sparsity of the attention map~\citep{beltagy2020longformer}, others express self-attention as a linear dot-product of kernel feature maps $\phi(\cdot)$~\citep{katharopoulos2020transformers}, or utilize random feature vectors~\citep{choromanski2020performer}. Proposed approaches reduce the computational complexity to $\mathcal{O}(N)$\footnote{In contrast, for \textit{Linear Transformer}~\citep{katharopoulos2020transformers, choromanski2020performer}, the complexity of linear self-attention is $\mathcal{O}(ND^2)$. In long sentences, $N$ is assumed to be around thousands of tokens. Therefore, switching to linear self-attention appears beneficial.}. One of the promising variants of a transformer is the \textit{Linear Transformer}~\citep{katharopoulos2020transformers} since, along with linear complexity, it requires constant $\mathcal{O}(1)$ memory in auto-regressive language modeling. Experiments with the long sequence benchmark Long Range Arena (LRA)~\citep{tay2020long}\footnote{In benchmark sequences ranging from $1$K to $16$K tokens} have indeed shown that the \textit{Linear Transformer} is $5$x times faster than the vanilla Transformer in training speed. However, the drawback of this architecture is lower performance compared to the original Transformer.

One way to reduce the performance gap between the \textit{Linear Transformer} and the original one is to select a more suitable kernel function $\phi(\cdot)$ in linear attention ~\citep{choromanski2020performer, schlag2021linear}. The poor performance of efficient transformers on LRA can also be attributed to the model's ability to capture positional information. The original Transformer model utilizes only absolute positional information, which is added through positional embeddings to contextual embeddings of the tokens. Other approaches, which enrich self-attention with additional information about relative distances between tokens, have recently shown visible improvements in performance. Some of them directly add a matrix of relative distances to the attention map~\citep{shaw2018self}, others compute separate attention scores between positional embeddings~\citep{he2020deberta}. We hypothesize that adding relative positional information could improve efficient transformers. However, most of the current implementations possess quadratic computational complexity, which neutralizes all efficiency of the \textit{Linear Transformer}. To deal with this problem, a linear complexity stochastic positional encoding (SPE) was proposed~\citep{liutkus2021relative}. Despite linear asymptotic, SPE remains relatively inefficient in training time due to its stochastic nature, while the improvement in accuracy it brings is relatively small on several LRA tasks. 

The contribution of this paper is two-fold. At first, we propose the \textbf{S}hift-\textbf{I}nvariant \textbf{K}ernel \textbf{F}unction (\textbf{SIKF}). It could be used as a kernel for the \textit{Linear Transformer} model and holds the shift-invariance property of $\softmax$ in the original attention. Second, we propose \textbf{F}ast \textbf{R}elative \textbf{P}ositional \textbf{B}ias (\textbf{FastRPB}) --- a Fast Fourier Transform-based bias for self-attention that represents relative positional information within sequences, has $\mathcal{O}(N\log N)$ complexity and requires only $\mathcal{O}(N)$ memory. FastRPB is orthogonal to the self-attention mechanism and can be combined with both efficient and original implementations.

We observed that SIKF is comparable to more complex kernels~\citep{choromanski2020performer, schlag2021linear} while being as fast as the original one~\citep{katharopoulos2020transformers}. We also evaluated FastRPB under different long-context scenarios, such as image classification and Long Range Arena tasks. Through a comprehensive study, we showed that the proposed technique outperforms the prior fast positional encoding method~\citep{liutkus2021relative} by a significant margin without adding a substantial computational footprint.

\begin{figure}
\centering
\begin{floatrow}[2]
  \centering
  \ffigbox{
    \caption{Learned weights $w_i$ assigned to pairwise distances between tokens $i$ in FastRPB 1D for different text LRA tasks.}
    \label{fig:sub1}
}{
    \centering
    \includegraphics[width=1.\linewidth]{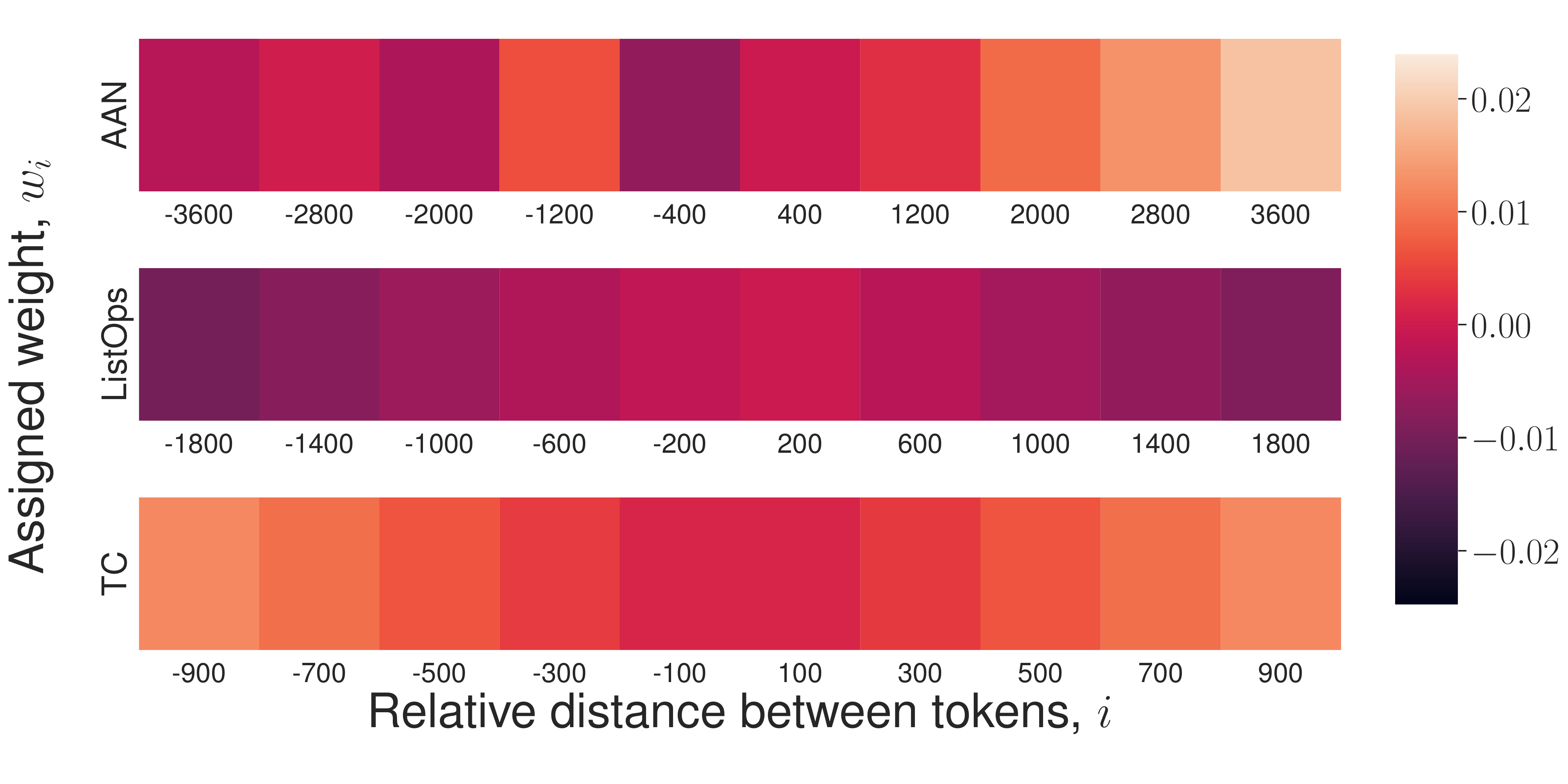}
  }
  \ffigbox{
    \caption{Learned FastRPB 2D weights assigned to distances from pixel (12, 10) to each other pixel in MNIST $28 \times 28$ image classification.}
    \label{fig:sub2}
}{
    \centering
    \includegraphics[width=1.\linewidth]{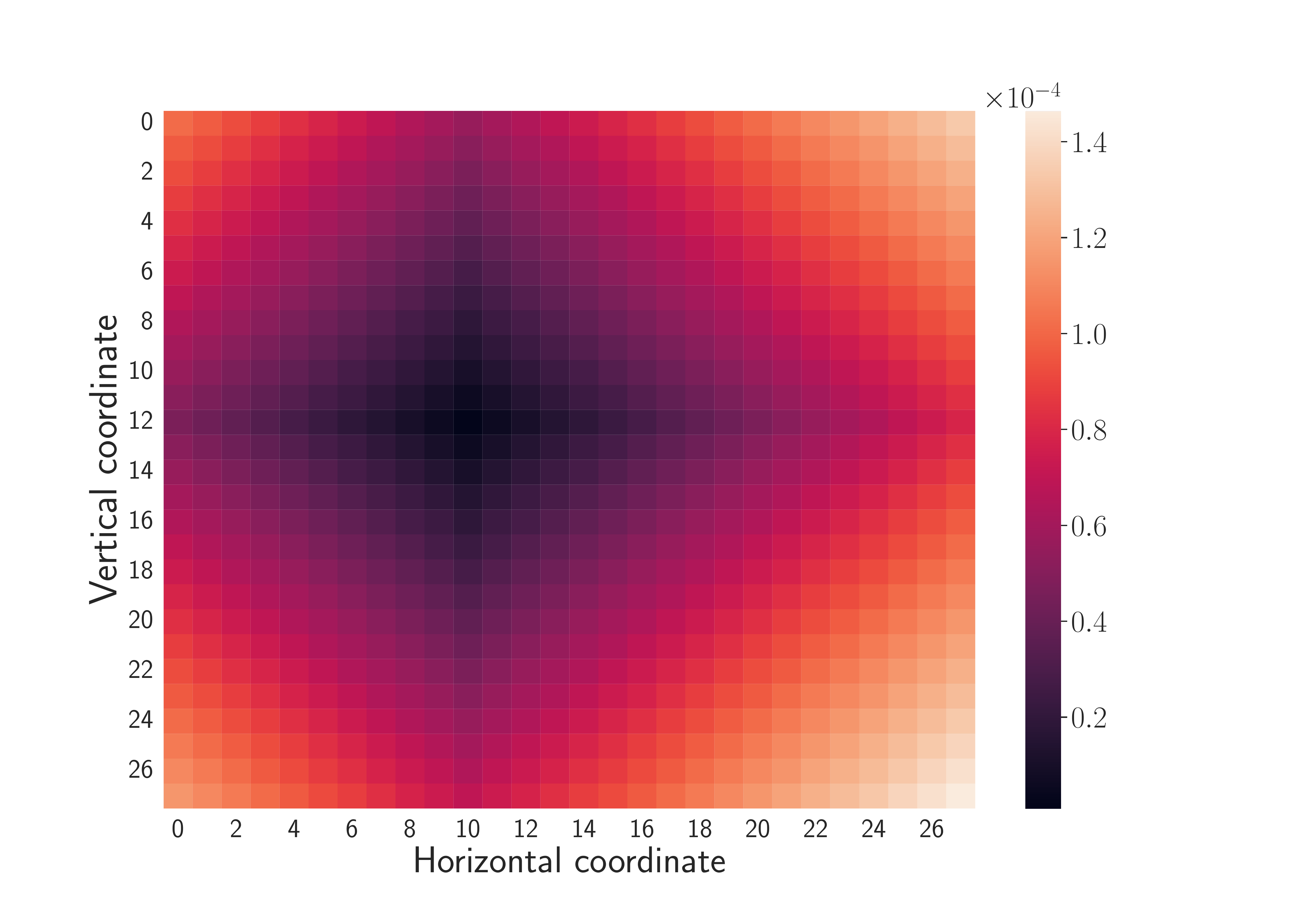}
  }
\end{floatrow}
\end{figure}
\section{Related Work}

\subsection{Attention Mechanism}

The core component of the Transformer~\citep{vaswani2017attention} is the attention layer, which computes attention weights $A_{m,n}$ that measure how important the role of the $n$-th key word is in shaping the meaning of the $m$-th output word. Using $A_{m,n}$, we can construct an attention matrix $\mA \in \mathbb{R}^{M \times N}$, and rewrite the equation using a matrix notation. The output of the attention layer $\mY$ is defined based on three matrices $\mQ \in \mathbb{R}^{M \times D}, \mK \in \mathbb{R}^{N \times D}$ and $\mV \in \mathbb{R}^{N \times D}$ (Queries, Keys, and Values) as follows:
\begin{align}
    \mY = \mA\mV = \softmax(\bm{\mathcal{A}})\mV  = \softmax(\mQ \mK^T / \sqrt{D})\mV
\label{formula:orig-attn}
\end{align}
In the vanilla Transformer, the attention matrix $\mA$ is computed explicitly, which leads to a $\mathcal{O}(MND)$ complexity, and $\mathcal{O}(MN)$ memory to store the matrix\footnote{In case of self-attention, $M$ equals $N$, and thus the complexity is $\mathcal{O}(N^2D)$ and memory requirement is $\mathcal{O}(N^2)$.}.

\begin{figure}
\centering
\begin{subfigure}{.49\textwidth}
  \centering
  \includegraphics[width=1.\linewidth]{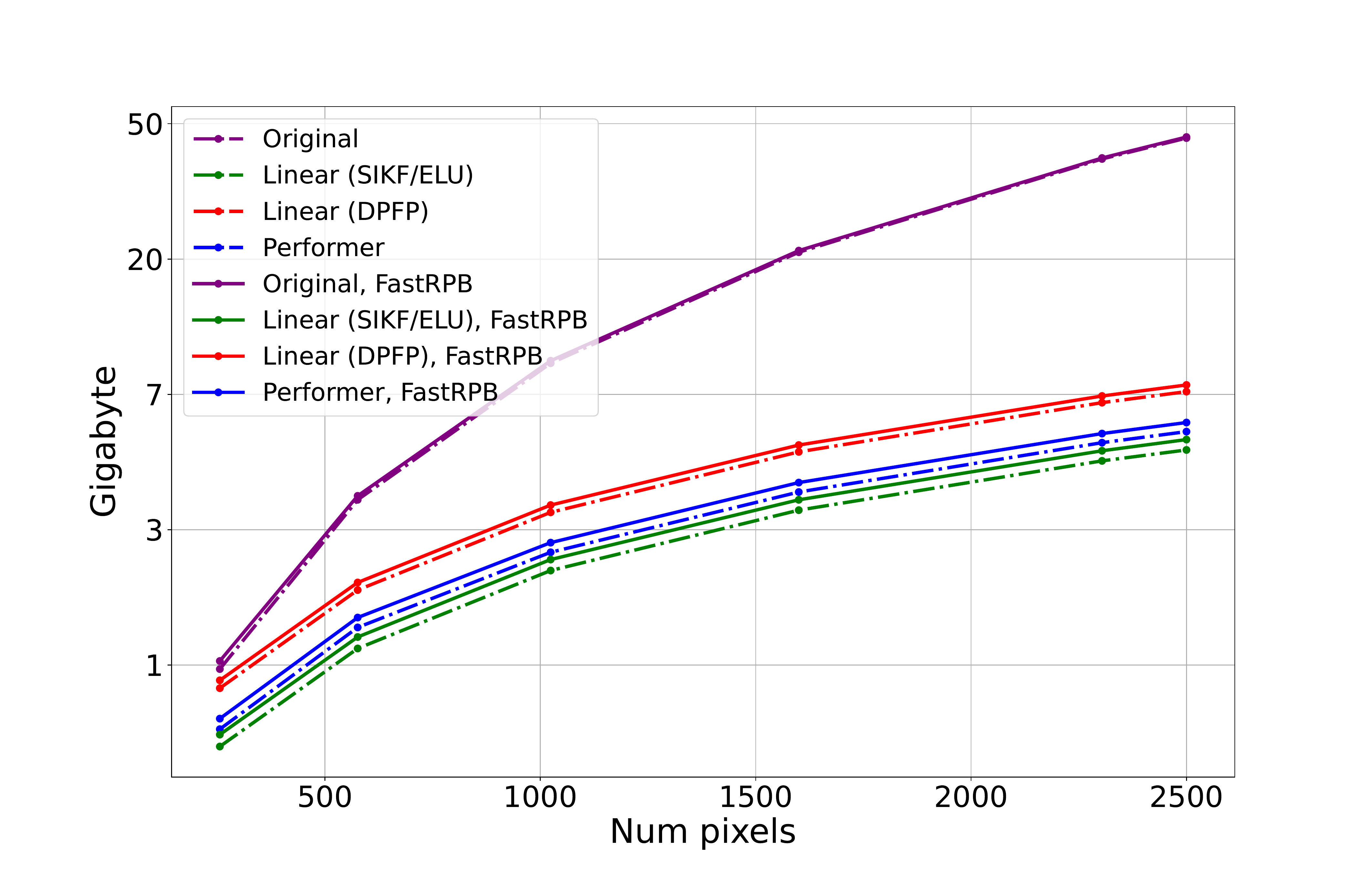}
  \caption{Evaluation memory consumption.}
  \label{fig:mem}
\end{subfigure}
\hspace{1mm}
\begin{subfigure}{.49\textwidth}
  \centering
  \includegraphics[width=1.\linewidth]{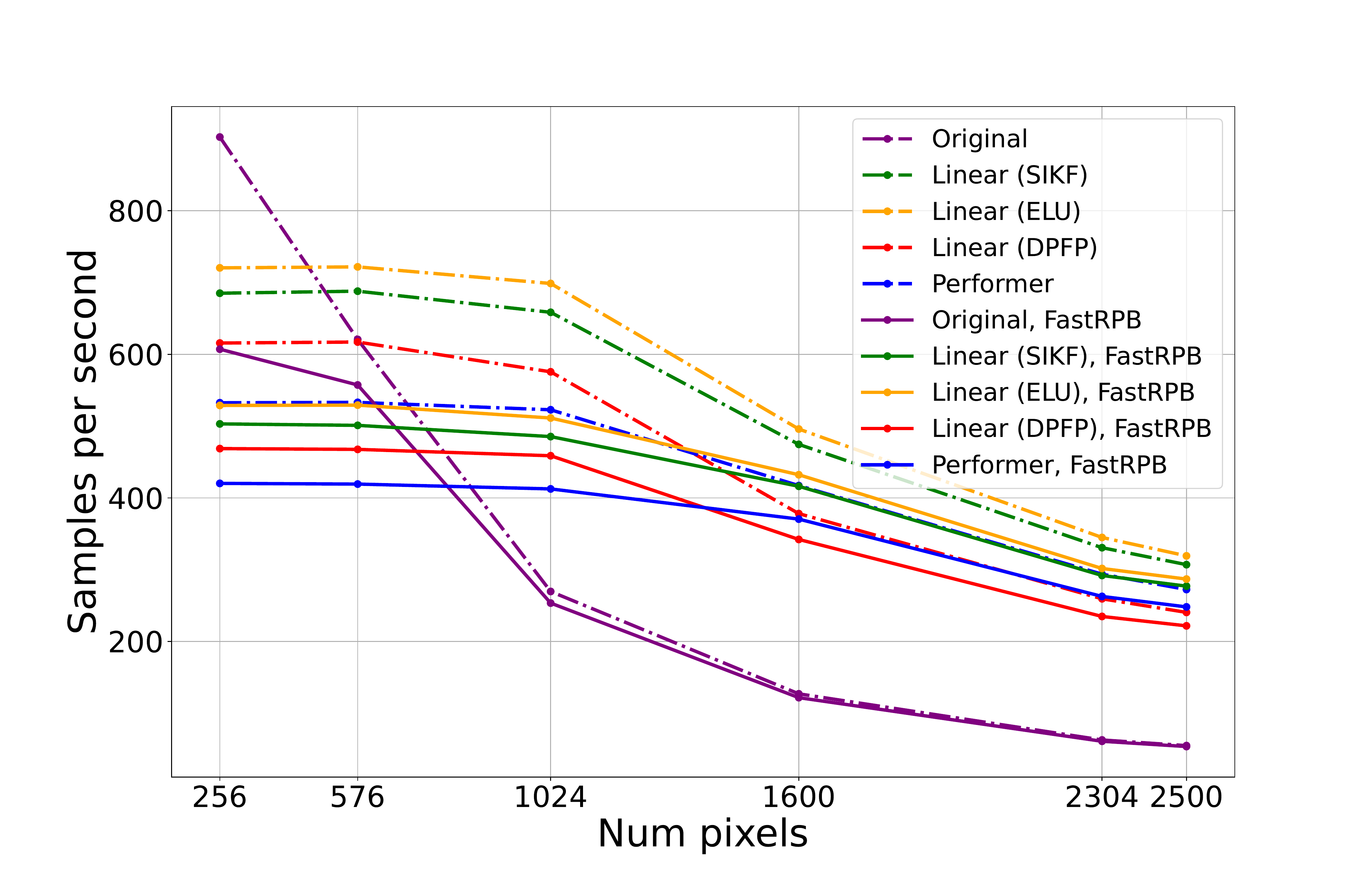}
  \caption{Evaluation time.}
  \label{fig:time}
\end{subfigure}
\caption{Evaluation time and memory for various types of Transformers on Nvidia A100 with respect to the number of pixels in the input image. To account for memory consumption, the y-axis is log-scaled.}
\label{fig:test}
\end{figure}

\subsection{Efficient Attention Mechanism}
Variants of the \textit{Linear Transformer}~\citep{katharopoulos2020transformers, choromanski2020performer} are a way to reduce the complexity of attention from quadratic to linear using the associative property of matrix products and kernel reformulation of attention. 

By substituting the $\softmax$ function in Equation~\ref{formula:orig-attn}, we obtain the $m$-th row $\vy_m$ of the matrix $\mY$:
\begin{align}
    \vy_{m} = 
    \frac{\sum_{n} \exp(\vq_m^T\vk_n / \sqrt{D}) \vv_n}{\sum_{n} \exp(\vq_m^T\vk_n / \sqrt{D})} =
    \frac{\sum_{n} \text{sim}(\vq_m, \vk_n) \vv_n}{\sum_{n} \text{sim}(\vq_m, \vk_n)}
\label{formula:general-attn-raw}
\end{align}
where $\exp(\vq_m^T \vk_n / \sqrt{D})$ is generalized by any arbitrary defined similarity function $\text{sim}(\vq_m, \vk_n)$.

The core idea of the \textit{Linear Transformer} is to replace $\text{sim}(\vq_m, \vk_n)$ with a dot-product using a kernel function $\phi(\cdot)$ and then use an associative property of matrix products as follows:
\begin{align}
    \vy_{m} = \frac{\sum_{n} \phi(\vq_m)^T \phi(\vk_n) \vv_n}{\sum_{n} \phi(\vq_m)^T\phi(\vk_n)} =  \frac{\phi(\vq_m)^T \sum_{n} \phi(\vk_n) \vv_n}{\phi(\vq_m)^T \sum_{n} \phi(\vk_n)}
\label{formula:linear-attn-raw}
\end{align}
The original attention mechanism has $\mathcal{O}(N^2D)$ time complexity, where $N$ represents the sequence length, and $\mathcal{O}(N^2)$ the memory footprint. While linear attention has the time and memory complexity of $\mathcal{O}(ND^2)$, which scales linearly with the sequence length $N$.

\subsection{Kernel Function Variants}
\label{kernels}

Selecting an appropriate kernel function for the \textit{Linear Transformer} remains an open question since different kernel functions can have a dramatic effect on trained model accuracy and speed.

\textbf{ELU + 1}. The originally proposed kernel is an element-wise $\text{ELU}(\cdot) + 1$~\citep{katharopoulos2020transformers}:
\begin{align}
    \phi(x) = \text{ELU}(x) + 1 = \begin{cases}x+1,& x>0 \\ \exp(x), &x\leq 0\end{cases}
\end{align}
The choice of $\text{ELU}(\cdot) + 1$ over $\text{ReLU}(\cdot)$ was prompted by its non-zero gradients for negative values.

\textbf{Performer}. The core idea is to approximate the $\softmax$ on average using random features~\citep{choromanski2020performer}. The kernel function is evaluated as:
\begin{align}
     \phi(\vx) = \frac{h(\vx)}{\sqrt{m}} \begin{bmatrix} \exp(\rmR\vx) \\ \exp(-\rmR\vx)\end{bmatrix}, \text{ where} \ h(\vx) = \frac{1}{\sqrt{2}}\exp\left(- \frac{1}{2}\|\vx\|\right)
\end{align}
Here $\begin{bmatrix} \exp(\rmR\vx) \\ \exp(-\rmR\vx)\end{bmatrix}$ stands for a concatenation of vectors $\exp(\rmR\vx)$ and $\exp(-\rmR\vx)$ along the feature dimension, each row $\rvr \in \mathbb{R}^{D}$ of matrix $\rmR \in \mathbb{R}^{R \times D}$ is sampled from normal distribution $\mathcal{N}(\bm{0}, \mI_{D})$, and dimension size $R$ is a hyperparameter. 

The main drawback of the $\text{Performer}$ is that the sampling of matrix $\rmR$ requires extra computations and introduces variance into the model’s output.

\textbf{DPFP}. Deterministic parameter-free projection is an alternative approach~\citep{schlag2021linear}. The kernel function, designed to facilitate orthogonality in the projected space $\mathbb{R}^{D_{\text{proj}}}$, is described as follows:
\begin{align}
    \phi_{i \cdot \nu}(\vx) = \text{ReLU}\left( \begin{bmatrix} \vx \\ -\vx\end{bmatrix}\right)_i\text{ReLU}\left( \begin{bmatrix} \vx \\ -\vx\end{bmatrix}\right)_{i+\nu} , \text{ where} \ \phi : \mathbb{R}^{D} \rightarrow \mathbb{R}^{D_{\text{proj}}}
\end{align}
here $i \cdot \nu$ indicates the index of vector $\phi(\vx)$, $i \in \{1, 2, ..., 2D\}$ is an index and $\nu \in \{1, 2, ..., 2D - 1\}$ is a hyperparameter that controls the capacity of the kernel function $\phi(\cdot)$. The \textit{Linear Transformer} with the $\text{DPFP}$ model outperforms models with a default kernel and Performer, even if $D_{\text{proj}}$ is relatively small. In addition, $\text{DPFP}$ showed speeds faster than models utilizing random features, but still slightly slower than $\text{ELU} + 1$.

\subsection{Positional Information}

Attention is permutation-invariant, which means that the attention layer does not make use of the sequence order. There exist different ways to encode positional information in the attention mechanism:

\textbf{Absolute Positional Encoding (APE)}, proposed in the original Transformer architecture, uses real-valued vector $\vp_i \in \mathbb{R}^{D}$ assigned to each position $i$. Some approaches, such as the vanilla Transformer~\citep{vaswani2017attention}, use predefined vectors, while others employ learnable vectors, e.g., in BERT~\citep{devlin2018bert}.

\textbf{Relative Positional Encoding (RPE)} is complement to the absolute positional encoding, which explicitly adds relative positional information between vectors~\citep{shaw2018self} to the model. ~\citet{raffel2019exploring} proposed to directly embed positional information into the matrix $\bm{\mathcal{A}}$ (see Equation ~\ref{formula:orig-attn}). This approach was then improved by separating the semantic correlation of words and their positional correlation by~\citet{ke2020rethinking}. The component $\mathcal{A}_{m,n}$ of matrix $\bm{\mathcal{A}}$ was then calculated as follows:
\begin{align}
    \mathcal{A}_{m,n} = \frac{1}{\sqrt{D}} \vq_m^T \vk_n  + \frac{1}{\sqrt{D}} (\mU_Q\vp_m)^T(\mU_K\vp_n)
\end{align}
where  $\vp_n$ and $\vp_m$ are the embeddings of the corresponding positions $n$ and $m$, and $\mU_Q, \mU_K \in \mathbb{R}^{D \times D}$ are learnable projection matrices for the positional embedding. 

By design, these approaches have quadratic computational complexity, which makes their usage with \textit{Linear Transformer} challenging since the naive application will neutralize all effectiveness of linear computation time.

To the best of our knowledge, \textbf{Stochastic Positional Encoding (SPE)}, proposed by~\citet{liutkus2021relative}, is currently the only positional encoding method compatible with \textit{Linear Transformer} variants due to its linear complexity. The key idea for SPE is to represent the attention relative distances matrix as a covariance. Following the notation from equation~\ref{formula:orig-attn}, we can express $\mathcal{A}_{m,n}$ as follows:
\begin{align}
    \mathcal{A}_{m,n} = \sum_{d=1}^D Q_{m, d} \cdot \mathcal{P}_{d}(m, n) \cdot K_{n, d} / \sqrt{D}, \text{ where } \mathcal{P}_{d}(m, n)=\mathbb{E}\big[\overline{\rq}_{d}(m) \cdot \overline{\rk}_{d}(n)\big]
\end{align}
where $Q_{m, d}$ and $K_{n, d}$ are components of matrices $\mQ$ and $\mK$ respectively. $\overline{\rq}_{d}(m)$ and $\overline{\rk}_{d}(n)$ are two real and zero-mean random variables, such that their covariance function matches $\mathcal{P}_d$. Varying the structure of matrices, $\mathcal{P}_d$ authors designed two variants of SPE: sinSPE and convSPE. The first one yields periodic covariance functions, which showed to be beneficial in tasks such as music generation. The second utilizes vanishing covariance functions, a promising concept introduced in ~\citet{wang2020encoding}, which yields notably smaller validation losses in some SPE experiments.

Although SPE was beneficial in some music generation tasks, it still requires many computations due to its stochastic nature. In practice, it could be dozens of times slower than the original Transformer, as we will show further.

\begin{table}[]
    \centering
    \small\addtolength{\tabcolsep}{-2.4pt}
\begin{tabular}{llccccc}
    \toprule
        & & \multicolumn{5}{c}{Positional Encoding} \\
        \cmidrule(l){3-7}
        & & None            & FastRPB        & sinSPE           & convSPE & RPE\\
   \midrule
\multicolumn{1}{l|}{\multirow{5}{*}{\rotatebox[origin=c]{90}{\parbox[c]{1cm}{\centering AAN}}}} &  \multicolumn{1}{l|}{Original} & OOM & OOM & OOM & OOM & OOM \\
\multicolumn{1}{l|}{} &  \multicolumn{1}{l|}{Linear, DPFP} & 61.01 ± 0.79 & 64.79 ± 1.52 & 61.53 ± 0.75 & 63.52 ± 0.71 & N/A \\
\multicolumn{1}{l|}{} &  \multicolumn{1}{l|}{Linear, SIKF} & 59.51 ± 0.3 & \textbf{67.19 ± 1.64} & 62.0 ± 0.36 & 58.93 ± 1.65 & N/A \\
\multicolumn{1}{l|}{} &  \multicolumn{1}{l|}{Linear, ReLU} & 58.78 ± 0.93 & \uuline{64.94 ± 1.6} & 62.39 ± 0.59 & 61.00 ± 1.34 & N/A \\
\multicolumn{1}{l|}{} &  \multicolumn{1}{l|}{Performer} & 59.84 ± 1.46 & 66.65 ± 0.91 & 60.00 ± 1.20  & 57.22 & N/A \\
\midrule
\multicolumn{1}{l|}{\multirow{5}{*}{\rotatebox[origin=c]{90}{\parbox[c]{1cm}{\centering ListOps}}}} &  \multicolumn{1}{l|}{Original} & 14.43 ± 4.73 & 14.6 ± 4.14 & -- & -- & OOM \\
\multicolumn{1}{l|}{} & \multicolumn{1}{l|}{Linear, DPFP} & \textbf{20.67 ± 3.95} & \uuline{17.97 ± 11.68} & 17.57 ± 0.18 & 16.17 ± 5.89 & N/A \\ 
\multicolumn{1}{l|}{} &  \multicolumn{1}{l|}{Linear, SIKF} & 12.55 ± 3.8 & 11.47 ± 4.79 & 15.25 ± 8.97 & 17.8 ± 0.0 & N/A \\
\multicolumn{1}{l|}{} &  \multicolumn{1}{l|}{Linear, ReLU} & 17.58 ± 1.01 & 17.67 ± 0.59 & 17.80 ± 0.00 & 9.50 ± 1.17 & N/A \\
\multicolumn{1}{l|}{} &  \multicolumn{1}{l|}{Performer} & 17.80 ± 0.00 & 17.75 ± 0.39 & 17.43 ± 0.32 & 17.80 & N/A \\ 
\midrule
\multicolumn{1}{l|}{\multirow{5}{*}{\rotatebox[origin=c]{90}{\parbox[c]{1cm}{\centering CIFAR}}}}  & \multicolumn{1}{l|}{Original} & 41.88 ± 0.48 & 39.02 ± 0.22 & -- & -- & N/A \\
\multicolumn{1}{l|}{} &  \multicolumn{1}{l|}{Linear, DPFP} & 41.79 ± 0.27 & 38.73 ± 0.09 & 41.97 ± 1.24 & 41.33 ± 0.84 & N/A \\
\multicolumn{1}{l|}{} &  \multicolumn{1}{l|}{Linear, SIKF} & 41.96 ± 0.47 & 38.89 ± 0.15 & 40.73 ± 0.58 &  \textbf{42.94 ± 0.51} & N/A  \\
\multicolumn{1}{l|}{} &  \multicolumn{1}{l|}{Linear, ReLU} & \uuline{42.25 ± 0.01} & 38.44 ± 0.38 & 41.21 ± 1.18 & 39.96 ± 1.31 & N/A \\
\multicolumn{1}{l|}{} &  \multicolumn{1}{l|}{Performer} & 41.81 ± 1.16 & 32.26 ± 9.53 & 41.12 ± 1.70 & 40.06 & N/A \\
\midrule
\multicolumn{1}{l|}{\multirow{5}{*}{\rotatebox[origin=c]{90}{\parbox[c]{1cm}{\centering TC}}}} &  \multicolumn{1}{l|}{Original} & 62.27 ± 0.8 & 62.02 ± 2.02 & -- & -- & 55.7 ± 1.94 \\
\multicolumn{1}{l|}{} &  \multicolumn{1}{l|}{Linear, DPFP} & 62.78 ± 0.48 & 63.05 ± 0.62 & 62.76 ± 0.21 & 62.78 ± 0.48 & N/A\\
\multicolumn{1}{l|}{} &  \multicolumn{1}{l|}{Linear, SIKF} & 61.64 ± 0.82 & 62.35 ± 0.24 & \uuline{63.37 ± 1.4} & 62.24 ± 0.56 & N/A\\
\multicolumn{1}{l|}{} &  \multicolumn{1}{l|}{Linear, ReLU} & 58.78 ± 0.93 & \textbf{63.95 ± 0.16} & 62.39 ± 0.59 & 61.00 ± 1.34 & N/A \\
\multicolumn{1}{l|}{} &  \multicolumn{1}{l|}{Performer} & 59.84 ± 1.46 & 62.66 ± 0.11 & 60.00 ± 1.20 & 57.22 & N/A\\
      \bottomrule
    \end{tabular}
    \caption{Experiments on the Long Range Arena benchmark. The best model is in bold, the double underline denotes the second-best result. Results for Performer and \textit{Linear Transformer} (ReLU) are copied from SPE ~\citep{liutkus2021relative}, except for the experiments with FastRPB. We mark experiments that failed due to memory limitations as OOM (Out of Memory). Since RPE is compatible only with the Original Transformer, we marked other experiments as Not Applicable (N/A). RPE is N/A for CIFAR since plain RPE is designed for 1D sequences. We marked experiments that were too long to train as "--". }
    \label{tab:lra}
\end{table}
\section{Approach}

\subsection{Shift-invariant Kernel Function (SIKF)}

We hypothesize that the shift-invariance property of the $\softmax$ function (i.e., the fact that $\softmax_i(\vx + c) = \softmax_i(\vx)$, where $\vx$ is some vector and $c$ is a constant that is added to every component of $\vx$) is an important property that makes the original Transformer perform better than a \textit{Linear Transformer} with an arbitrary kernel. Based on this assumption, we propose SIKF as $\phi(x) = \exp{(x)}$, which satisfies the property of shift-invariance. If we substitute this function in the linear attention mechanism from Equation~\ref{formula:linear-attn-raw}, then for every real-valued constants $c$ and $d$ we will get:
\begin{align}
    \frac{\phi(\vq_m + c)^T\sum_{n} \phi(\vk_n + d)\vv_n}{\phi(\vq_m + c)^T\sum_{m} \phi(\vk_n + d)} = \frac{e^{c}\phi(\vq_m)^T \sum_{n} e^{d}\phi(\vk_n) \vv_n}{e^{c}\phi(\vq_m)^T \sum_{n} e^{d}\phi(\vk_n)} = \frac{\phi(\vq_m)^T \sum_{n} \phi(\vk_n) \vv_n}{\phi(\vq_m)^T \sum_{n} \phi(\vk_n)}
\end{align}
Thus, attention in the \textit{Linear Transformer} with $\exp(\cdot)$ kernel function holds the same shift-invariance property as plain $\softmax$.

Based on our experiments, we conclude that SIKF is faster than Performer and DPFP, at the same time having comparable accuracy. In addition, it does not provide an extra memory footprint, which is essential for scaling the \textit{Linear Transformer} on extremely long sequences.

\subsection{Fast Relative Positional Bias (FastRPB)}

Although adding positional information in the attention mechanism is beneficial for model accuracy, current approaches are relatively inefficient for long sequences in terms of speed and memory footprint. In this context, it is desirable to design an approach that will add relative positional information to attention efficiently, while simultaneously being compatible with various efficient attention modifications. To achieve this goal, we propose FastRPB\footnote{We though of naming FastRPB as FastRPE to represent that it is like a faster RPE, but changed one letter to emphasize that FastRPB is orthogonal to the selection of an attention algorithm and could be seen as a separate bias term to the attention map.} as a separate term for attention. 

The output matrix $\mY$ of an attention layer with FastRPB is defined as:
\begin{align}
    \mY = \text{AttentionVariant}(\mQ, \mK, \mV) + \mW \mV
\label{formula:pb-attn}
\end{align}
where matrix $\mW \in \mathbb{R}^{M \times N}$ consists of learnable weights $W_{m,n}$ representing relative distances between $m$ and $n$ embedding vectors from matrix $\mV$. Note that Equation \ref{formula:pb-attn} is invariant of choosing a specific attention mechanism and could be used with both vanilla attention and its linear variants (Equations~\ref{formula:orig-attn} and~\ref{formula:linear-attn-raw} respectively).

One can think of the matrix $\mW$ as a bias term to the usual attention matrix $\mA$ from equation~\ref{formula:orig-attn}, correcting the attention weights according to the relative distance between the corresponding tokens. However, adding a positional bias term in the Equation \ref{formula:pb-attn} still requires $\mathcal{O}(NMD)$ computations due to the matrix product and $\mathcal{O}(NM)$ memory to store the bias matrix $\mW$\footnote{$\mathcal{O}(N^2D)$ and $\mathcal{O}(N^2)$ respectively in the case of self-attention}. In this regard, in the following two subsections, we will construct the FastRPB positional bias terms matrices $\mW_{1d}$ and $\mW_{2d}$ for different types of sequences that can be efficiently multiplied by $\mV$. $\mW_{1d}$ will be utilized for 1D sequences (e.g., natural language texts), and its coefficients correspond to distances between words in 1D sequences. For 2D sequences, we will utilize $\mW_{2d}$, coefficients of which represent distances between elements of 2D sequences (i.e., pixels). We will show that these specific matrices $\mW_{1d}$ and $\mW_{2d}$ could be multiplied with $\mV$ using only $\mathcal{O}(DN\log{N})$ computations, and requiring only $\mathcal{O}(N)$ memory.

Further in this article, we work with self-attention --- a variant of the attention mechanism where input and output sequences lengths are the same, i.e. $N = M$. In the general case of the attention mechanism, when we have an input sequence of length $N$ and an output sequence of length $M$, we can pad the longer one to make the input and output lengths match.

\begin{table}[]
    \centering
    \small\addtolength{\tabcolsep}{-2.4pt}
\begin{tabular}{l|c|c}
    \toprule
    Model & w/o FastRPB & w/ FastRPB\\ 
       \midrule
    Original &  97.34 ± 0.23 & \textbf{98.27 ± 0.19}\\
    Linear, DPFP & 97.09 ± 0.19 & \uuline{97.66 ± 0.22}\\
    Linear, SIKF & 96.49 ± 0.20 & 97.37 ± 0.35\\
    Linear, ELU + 1 & 94.01 ± 0.31 & 96.71 ± 0.40\\
    Performer &  96.6 ± 0.29 & 97.52 ± 0.26\\
\bottomrule
\end{tabular}
    \caption{MNIST F1 score. All the experiments were run on 4 Nvidia Tesla T4.}
    \label{tab:mnist}
\end{table}

\subsubsection{1D Sequence Case}
\label{section:1d}

Suppose we have a 1D sequence with $N$ tokens. In such a sequence, there are exactly $2N-1$ relative distances between tokens\footnote{Relative distance from $m$-th token to $n$-th token is $m-n$, which can have both positive and negative values. In this regard, we have exactly $2N-1$ learnable parameters}. Let's assign a learnable parameter $w_{i} \in \mathbb{R}$ for each relative distance $i \in \{-N+1 , \ ... , -1, 0, 1, \ ... , N-1\}$. We then will obtain a set of parameters: 
\begin{align}
    \{w_{-N+1}, \ ..., \ w_{-1}, \ w_{0}, \ w_{1}, \ ..., \ w_{N-1}\}
\end{align}
Next, we will construct a matrix $\mW_{1d}$ using parameters $\{w_i\}_{i=-N+1}^{N-1}$. The basic intuition is to make $(n,m)$-th element of matrix $\mW_{1d}$ to be assigned to the relative distance between the $m$-th token to the $n$-th token, i.e., $w_{m-n}$. Therefore, the matrix $\mW_{1d}$ will have the following structure:
\begin{align}
    \mW_{1d} = \begin{pmatrix} w_{0} & w_{1} & w_{2} & \cdots & w_{N-1} \\ w_{-1} & w_{0} & w_{1} & \cdots & w_{N-2} \\ \vdots & \vdots & \vdots & \ddots \\ w_{-N+1} & w_{-N+2} & w_{-N+3} & \cdots & w_{0} \end{pmatrix}
\end{align}
\label{toeplitz_matrix}
By definition, $\mW_{1d}$ is a Toeplitz matrix~\citep{toeplitz2001gray}. A naive way to calculate the product $\mW_{1d} \cdot \mV$ requires $O(N^2D)$ computations in case of self-attention\footnote{Matrix $\mV$ has size $N \times D$, where $D$ is a hidden size}. It turns out that it can be efficiently multiplied by a matrix $\mV$ according to the following proposition:
\begin{restatable}[]{proposition}{Toeplitz}
    \label{prop:toeplitz}
    For every Toeplitz matrix $\mW_{1d} \in \mathbb{R}^{N \times N}$ and for every matrix $\mV \in \mathbb{R}^{N \times D}$, matrix product $\mW_{1d} \cdot \mV$ requires $\mathcal{O}(DN \log{N})$ operations and $\mathcal{O}(N)$ memory. Here $N$ is length of the input sequence.
\end{restatable}
Using proposition~\ref{prop:toeplitz}, we can claim that FastRPB for a 1D sequence will require $\mathcal{O}(DN\log{N})$ computational operations. For generating the Toeplitz matrix $\mW_{1d}$, we only need $\mathcal{O}(N)$ memory for storing parameters $\{w_i\}_{i=-N+1}^{N-1}$. For the proof and a more detailed explanation of proposed properties, see Appendix~\ref{proof1}.

\subsubsection{2D Sequence Case}
\label{section:2d}

In the case of 2D sequences (e.g., images), a similar matrix to $\mW_{1d}$ could be defined. We will call this matrix $\mW_{2d}$, and it will consist of learnable weights assigned to pairwise distances from each pixel of the image to the rest of the pixels. Here we will consider only the case of square images of size $N \times N$ \footnote{If we work with non-square images of size $N \times M$, we can simply pad them with zeros to make it square.}.

The natural way to process images of the size $N \times N$ in the Transformer model is to flatten them into a vector of size $N^2$. In this regard, a matrix of pairwise distances $\mW_{2d}$ needs to be of size $N^2 \times N^2$. For simplicity, we will present images as a $N \times N$ matrix, and $\mW_{2d}$ will be expressed as a tensor $\tW_{2d}$ of size $(N \times N) \times (N \times N)$, in which the $(n,m,l,k)$ component represents distance from pixel $(l,k)$ to pixel $(n,m)$.

We will assume that the distance between two pixels is a sum of the vertical and horizontal distances\footnote{If we consider two pixels $p_1 = (3, 2)$ and $p_2 = (0, 1)$ of image of size $4 \times 4$, the horizontal relative distance from pixel $p_2$ to pixel $p_1$ then will be $2-1$, and the vertical will be $3-0$}. In this regard, a tensor $\tW_{2d}$ can be decomposed on vertical and horizontal tensor terms $\tX$ and $\tY$ as $\tW_{2d} = \tX + \tY$, respectively. Similar to the 1D case, we will assign shared learnable parameters $\{w_{i}\}_{i=-N+1}^{N-1}$ for horizontal and vertical distances. To compute a matrix product of tensor $\mW_{2d}$ of size $N^2 \times N^2$ with matrix $\mV$ of size $N^2 \times D$, we will then simply flatten the tensors $\tX$ and $\tY$ to obtain matrices $\mX_{\text{flat}}$ and $\mY_{\text{flat}}$ of shape $N^2 \times N^2$, and compute $\mW_{2d}\mV$ as $\mX_{\text{flat}}\mV + \mY_{\text{flat}}\mV$.

It turns out that the structure of matrices $\mX_{flat}$ and $\mY_{flat}$ is very similar to Toeplitz matrices from the Section~\ref{section:1d}. In this regard, $\mW_{2d}$ can be efficiently multiplied by $\mV$ according to the following proposition:
\begin{restatable}[]{proposition}{Toeplitz}
    \label{prop:toeplitz_2d}
    Product of matrix $\mW_{2d} \in \mathbb{R}^{N^2 \times N^2}$ with matrix $\mV \in \mathbb{R}^{N^2\times D}$ using matrices $\mX_{\text{flat}}$ and $\mY_{\text{flat}}$ requires $\mathcal{O}(DN\log{N})$ and $\mathcal{O}(N)$ memory.
\end{restatable}
Using the above proposition~\ref{prop:toeplitz_2d}, we can conclude that FastRPB 2D will require $\mathcal{O}(DN\log{N})$ computational operations. It is not essential to store whole tensors $\tX$ and $\tY$ to compute the product, as we only need $\mathcal{O}(N)$ of memory for parameters $\{w_i\}_{i=-N+1}^{N-1}$ generating these tensors. (See Appendix \ref{proof2} for the proof).

\begin{table}[]
    \centering
    \small\addtolength{\tabcolsep}{-2.4pt}
\begin{tabular}{clccccc|ccccc}
    \toprule
    & & \multicolumn{5}{c|}{Training time (hours)} & \multicolumn{5}{c}{Peak Memory Usage (GB)} \\
    \cmidrule(l){3-12}
      &  & None & FastRPB & sinSPE & convSPE & RPE & None & FastRPB & sinSPE & convSPE & RPE\\
       \midrule
\multicolumn{1}{l|}{\multirow{5}{*}{\rotatebox[origin=c]{90}{\parbox[c]{1cm}{\centering {AAN}}}}}  & \multicolumn{1}{l|}{Original} & OOM & OOM & OOM & OOM & OOM & 5.81 & 6.03 & -- & -- & 9.69 \\
\multicolumn{1}{l|}{} & \multicolumn{1}{l|}{Linear, DPFP} & \textbf{0.36} & \uuline{0.43} & 2.45 & 15.41 & N/A & \textbf{0.31} & \underline{0.57} & 0.78 & 0.87 & N/A \\
\multicolumn{1}{l|}{} & \multicolumn{1}{l|}{Linear, SIKF} & \textbf{0.36} & \underline{0.45} & 1.22 & 9.12 & N/A & \textbf{0.31} & \underline{0.57} & 0.78 & 0.83 & N/A \\
\multicolumn{1}{l|}{} & \multicolumn{1}{l|}{Linear, ReLU} & \textbf{0.36} & \underline{0.45} & 1.26 & 9.13 & N/A & \textbf{0.31} & \underline{0.57} & 0.78 & 0.83 & N/A \\
\multicolumn{1}{l|}{} & \multicolumn{1}{l|}{Performer} & 0.6 & 0.79 & 1.6 & 10.52 & N/A & \uuline{0.54} & 0.68 & 0.77 & 0.87 & N/A \\
\midrule
\multicolumn{1}{l|}{\multirow{5}{*}{\rotatebox[origin=c]{90}{\parbox[c]{1cm}{\centering {ListOps}}}}} & \multicolumn{1}{l|}{Original} & 0.74 & 0.85 & -- & -- & OOM & 3.25 & 3.49 & -- & -- & 3.66 \\
\multicolumn{1}{l|}{} & \multicolumn{1}{l|}{Linear, DPFP} & \uuline{0.26} & 0.36 & 2.3 & 13.8 & N/A & \uuline{0.68} & \underline{0.85} & 1.32 & 1.33 & N/A \\
\multicolumn{1}{l|}{} & \multicolumn{1}{l|}{Linear, SIKF} & \textbf{0.24} & \underline{0.34} & 0.95 & 6.85 & N/A & \uuline{0.68} & \underline{0.85} & 1.32 & 1.33 & N/A \\
\multicolumn{1}{l|}{} & \multicolumn{1}{l|}{Linear, ReLU} & \textbf{0.24} & \underline{0.34} & 0.98 & 6.87 & N/A & \uuline{0.68} & \underline{0.85} & 1.32 & 1.33 & N/A \\
\multicolumn{1}{l|}{} & \multicolumn{1}{l|}{Performer}& 0.38 & 0.48 & 1.06 & 8.7 & N/A & \textbf{0.67} & 0.9 & 1.03 & 1.32 & N/A \\
\midrule
\multicolumn{1}{l|}{\multirow{5}{*}{\rotatebox[origin=c]{90}{\parbox[c]{1cm}{\centering {CIFAR}}}}} & \multicolumn{1}{l|}{Original}& \textbf{1.94} & \underline{1.97} & -- & -- & N/A & \textbf{12.36} & \uuline{12.39} & -- & -- & N/A \\
\multicolumn{1}{l|}{} & \multicolumn{1}{l|}{Linear, DPFP} & \textbf{1.94} & \underline{1.97} & 2.07 & 2.44 & N/A & \textbf{12.36} & \uuline{12.39} & 12.57 & 12.58 & N/A \\
\multicolumn{1}{l|}{} & \multicolumn{1}{l|}{Linear, SIKF} & \textbf{1.94} & \uuline{1.96} & 2.06 & 2.43 & N/A & \textbf{12.36} & \uuline{12.39} & 12.57 & 12.58 & N/A \\
\multicolumn{1}{l|}{} & \multicolumn{1}{l|}{Linear, ReLU} & \textbf{1.94} & \underline{1.97} & 2.06 & 2.44 & N/A & \textbf{12.36}  & \uuline{12.39} & 12.57 & 12.58 & N/A \\
\multicolumn{1}{l|}{} & \multicolumn{1}{l|}{Performer} & \textbf{1.94} & \uuline{1.96} & 2.07 & 2.44 & N/A & \textbf{12.36} & \uuline{12.39} & 12.57 & 12.58 & N/A \\
\midrule
\multicolumn{1}{l|}{\multirow{5}{*}{\rotatebox[origin=c]{90}{\parbox[c]{1cm}{\centering {TC}}}}} & \multicolumn{1}{l|}{Original} & 1.81 & 2.24 & -- & -- & 6.58 & 0.52 & 0.52 & -- & -- &  0.78 \\
\multicolumn{1}{l|}{} & \multicolumn{1}{l|}{Linear, DPFP} & \uuline{1.48} & 1.58 & 4.25 & 13.56 & N/A & \textbf{0.23} & \textbf{0.23} & \uuline{0.33} & 0.43 & N/A \\
\multicolumn{1}{l|}{} & \multicolumn{1}{l|}{Linear, SIKF} & \underline{1.56} & 1.62 & 3.52 & 9.85 & N/A & \textbf{0.23} & \textbf{0.23} & \uuline{0.33} & 0.43 & N/A \\
\multicolumn{1}{l|}{} & \multicolumn{1}{l|}{Linear, ReLU} & \textbf{1.46} & 1.62 & 3.54 & 9.85 & N/A & \textbf{0.23} & \textbf{0.23} & \uuline{0.33} & 0.43 & N/A \\
\multicolumn{1}{l|}{} & \multicolumn{1}{l|}{Performer} & 1.93 & 2.38 & 5.81 & 40.09 & N/A & \uuline{0.33} & \underline{0.41} & \uuline{0.33} & 0.47 & N/A \\

     \bottomrule
     
\end{tabular}
    \caption{Benchmark results on LRA with experiment setup proposed in SPE ~\citep{liutkus2021relative}. All of the above experiments were conducted using a single Nvidia A100 GPU. The best model is in bold, the double underline denotes the second-best result, and the single underline indicates a third-best result. We mark experiments that failed due to limited memory as OOM (Out of Memory). We did not run Original Transformer with sinSPE and convSPE since they require too much time to train.}
    \label{tab:efficiency}
\end{table}

\section{Experiments}
\textbf{Long Range Arena}.
We evaluate proposed methods in the Long Range Arena~\citep{tay2020long}, a benchmark for efficient Transformers with several text and image long-sequence tasks. The main challenge of these tasks is dictated by the large sequence lengths, which average number of tokens can vary from $1K$ to $16K$\footnote{We did not include another synthetic image classification task, Pathfinder, since we were unable to reproduce the results obtained in the original paper~\citep{tay2020long}.}. In our experiments, we used the following tasks from this benchmark: (1) ListOps, which tests if a model is capable of parsing hierarchical expressions~\citep{nangia2018listops}; (2) TC, which consists of movie review sentiment analysis on the IMDB corpus~\citep{maas-EtAl:2011:ACL-HLT2011}; (3) All About NLP (AAN), which evaluates the model performance in matching and retrieval tasks~\citep{radev2013acl}; and (4) CIFAR10, an image classification dataset~\citep{Krizhevsky2009LearningML}.

We compared the vanilla Transformer and the \textit{Linear Transformer} with all kernels observed in section~\ref{kernels} with SIKF, combined with different positional encodings, namely sineSPE, convSPE, FastRPB. We also reported the results of our experiments without adding any relative positional information. All models used trainable Absolute Positional Encodings.

All experiments and hyperparameters were conducted following instructions for the LRA dataset. We also used LRA tasks to measure memory usage during evaluation and the computational footprint during the training.

\textbf{MNIST}. Due to the fact that in LRA CIFAR10 experiment only a single-layer transformer is used, we conducted another image recognition experiment with larger neural networks. We evaluated all the above models with and without FastRPB on the classical image classification dataset MNIST~\citep{726791}. In this experiment, we did not compare FastPRE with other positional encoding methods since, as we observed in LRA, they require dozens of times more training time in experiments with multi-layer transformers with a large hidden state size. 

For all experiments, we used a model with $8$ layers, $8$ attention heads, a hidden size equal to $256$, and batch equal to $160$. We trained models using AdamW optimizer and made 20 runs of the Bayesian hyperparameter search to find the optimal learning rate, and than trained all models for $25$ epochs. Parameters are presented in Appendix~\ref{sweep-lr}. 
We linearly decayed the learning rate to $0$ during the training. Final results are averaged over 10 runs with different random seed values.

\section{Results}
\textbf{Long Range Arena}. See Table~\ref{tab:lra} for the evaluation results. The \textit{Linear Transformer} with SIKF kernel comes out in the top two results for every dataset except for CIFAR10, which we will discuss separately. The memory footprint (see Table~\ref{tab:efficiency}) of SIKF is very close to the ReLU, DPFP, and Performer, while the DPFP and Performer consistently performed slower (up to \textbf{1.4x} times).

The \textit{Linear Transformer} using FastRPB showed significantly higher results on ANN and TC both in memory, speed, and accuracy, achieving even better results than the original Transformer. Moreover, architectures with FastRPB confirmed the above propositions~\ref{prop:toeplitz} and~\ref{prop:toeplitz_2d} by proving to have memory and computation consumption comparable with the default architectures. For the ListOps dataset, the best performance was obtained by the model without any relative positional encoding. We attribute this result to the fact that relative distances can be confusing in sparse hierarchical structure, such as expressions for ListOps or source code (e.g., the distance between \textsc{if} and \textsc{else} in source code can be pretty large, however, these statements are inwardly connected). We measured the memory usage and computational footprint of models (see Table \ref{tab:efficiency}), according to which FastRPB requires up to \textbf{30x} less time to train than convSPE, and up to \textbf{3x} less then sinSPE. Simultaneously, FastRPB has a \textbf{1.5x} smaller memory footprint on evaluation then sinSPE and convSPE. Therefore, we can conclude that FastRPB is the fastest and the most accurate method compared to the others. 

The sequence of FastRPB's learned weights for different text tasks is presented in Figure~\ref{fig:sub1}, where the $x$-axis represents the relative distance $m-n$ from $n$-th token to $m$-th token, and the color denotes the value of $(W_{1d})_{n, m} = w_{m-n}$. In the AAN task, FastRPB forces the model to attend more to the very end of the text. Such an observation can be attributed to the fact that AAN mainly consists of scientific texts, in which the conclusion can usually be found at the end. As for ListOps, learned FastRPB weights are usually relatively small, which supports the hypothesis that relative positional encodings in such tasks should be designed using the text hierarchy information. In the TC task, learned weights mainly draw the model's focus forward and backward to enable the model to capture long-range dependencies.

In experiments with CIFAR, usage of FastRPB was shown to decrease the model's performance, and the best result was obtained with convSPE, which outperformed others by a significant margin. We attributed this to the experiment setup using a single-layer network. In this regard, we conducted the experiments with a more extensive network on the MNIST dataset.

\textbf{MNIST}. In this task, each of the above models using FastRPB showed superior performance compared to the others (see Table \ref{tab:mnist}) while requiring a rather small amount of additional memory and computational time (see Figures~\ref{fig:mem} and~\ref{fig:time} respectively). As can be observed from the plots, that requirement for the \textit{Linear Transformer} with FastRPB is up to \textbf{10x} times less than that of the original one. Moreover, in terms of speed, the \textit{Linear Transformer} with FastRPB is $\textbf{5x}$ times faster in evaluation time compared to original Transformer. 

A detailed overview of the trained FastRPB can be found in Figure~\ref{fig:sub2}, where we learned the slice $(\tW_{2d})_{:, :, 12, 10}$ which represents pairwise distances from pixel $(12, 10)$ to every other pixel of $28 \times 28$ MNIST image. We observed that FastRPB forced the model to look at more distinct pixels rather than close ones.

\section{Conclusion}

We presented two novel approaches aimed at increasing the accuracy of the \textit{Linear Transformer} model without an additional memory footprint and significant loss in speed. The contribution of this paper is two-fold: we first make linear attention shift-invariant, and then add a bias term to attention scores, representing pairwise distances between tokens of the sequence. We computed this bias term efficiently and achieved $\mathcal{O}(N \log{N})$ complexity and $\mathcal{O}(N)$ memory w.r.t. sequence length.

We demonstrate the benefits of our approach compared to others on four long-sequence tasks from the Long Range Arena benchmark as well as on the MNIST dataset. Our model performs significantly better than previous approaches, obtaining the best accuracy on several tasks while being almost as efficient in terms of speed and memory consumption as the plain \textit{Linear Transformer}.

We believe that the principles presented in this work can serve as a basis for future research on the role of positional information encoding in transformer architectures. To this end, we make all the code and trained models open-source.

\bibliography{main}
\bibliographystyle{iclr2022_conference}

\appendix

\label{appendix}

\section{MNIST Hyperparameter Search}
\label{sweep-lr}
\begin{table}[hp]
    \centering
    \small\addtolength{\tabcolsep}{-2.4pt}
\begin{tabular}{l|c|c}
    \toprule
    Model & w/o FastRPB & w/ FastRPB\\ 
       \midrule
    Original & $1.05 \cdot 10^{-4}$ & $1.35 \cdot 10^{-4}$ \\
    Linear, DPFP & $1.3 \cdot 10^{-4}$ & $0.7 \cdot 10^{-4}$ \\
    Linear, SIKF & $1.25 \cdot 10^{-4}$  & $1.45 \cdot 10^{-4}$ \\
    Linear, ELU + 1 & $1.3 \cdot 10^{-4}$ & $1.25 \cdot 10^{-4}$ \\
    Performer &  $1.2 \cdot 10^{-4}$ & $1.0 \cdot 10^{-4}$ \\
\bottomrule
\end{tabular}
    \caption{Best Learning Rate values obtained from 20 runs of Bayesian hyperparameter search}
    \label{tab:mnist-params}
\end{table}

\begin{figure}[hp]
\begin{subfigure}{.8\textwidth}
  \centering
  \includegraphics[width=1.\linewidth]{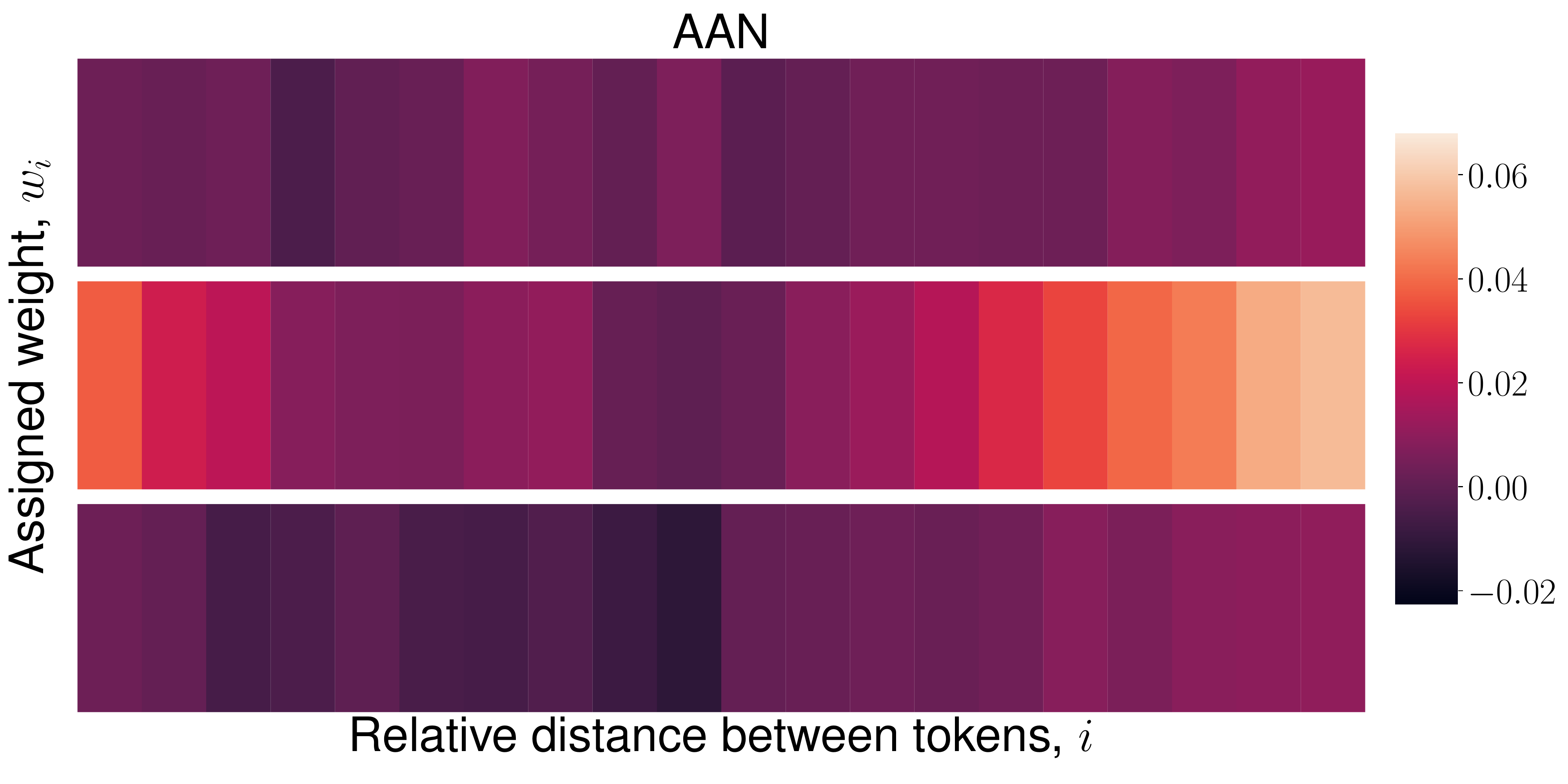} 
  \caption{Weights $w_i$, learned on the AAN task, assigned to the pairwise distances between tokens $i$ in FastRPB 1D.}
    \label{fig:aan}
\end{subfigure}
\begin{subfigure}{.8\textwidth}
  \centering
  \includegraphics[width=1.\linewidth]{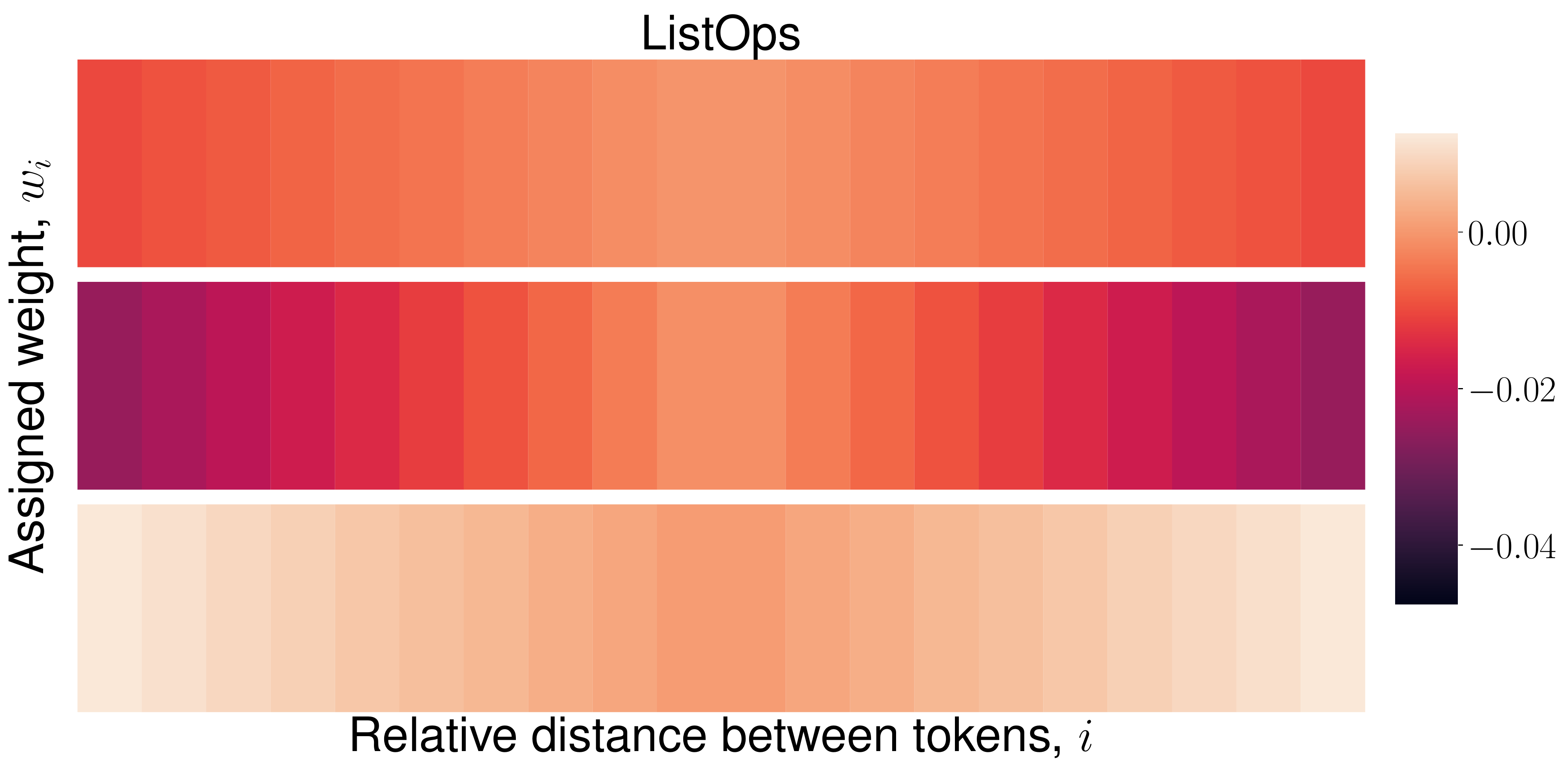}
  \caption{Weights $w_i$, learned on the ListOps task, assigned to the pairwise distances between tokens $i$ in FastRPB 1D.}
    \label{fig:listops}
\end{subfigure}
\begin{subfigure}{.8\textwidth}
  \centering
  \includegraphics[width=1.\linewidth]{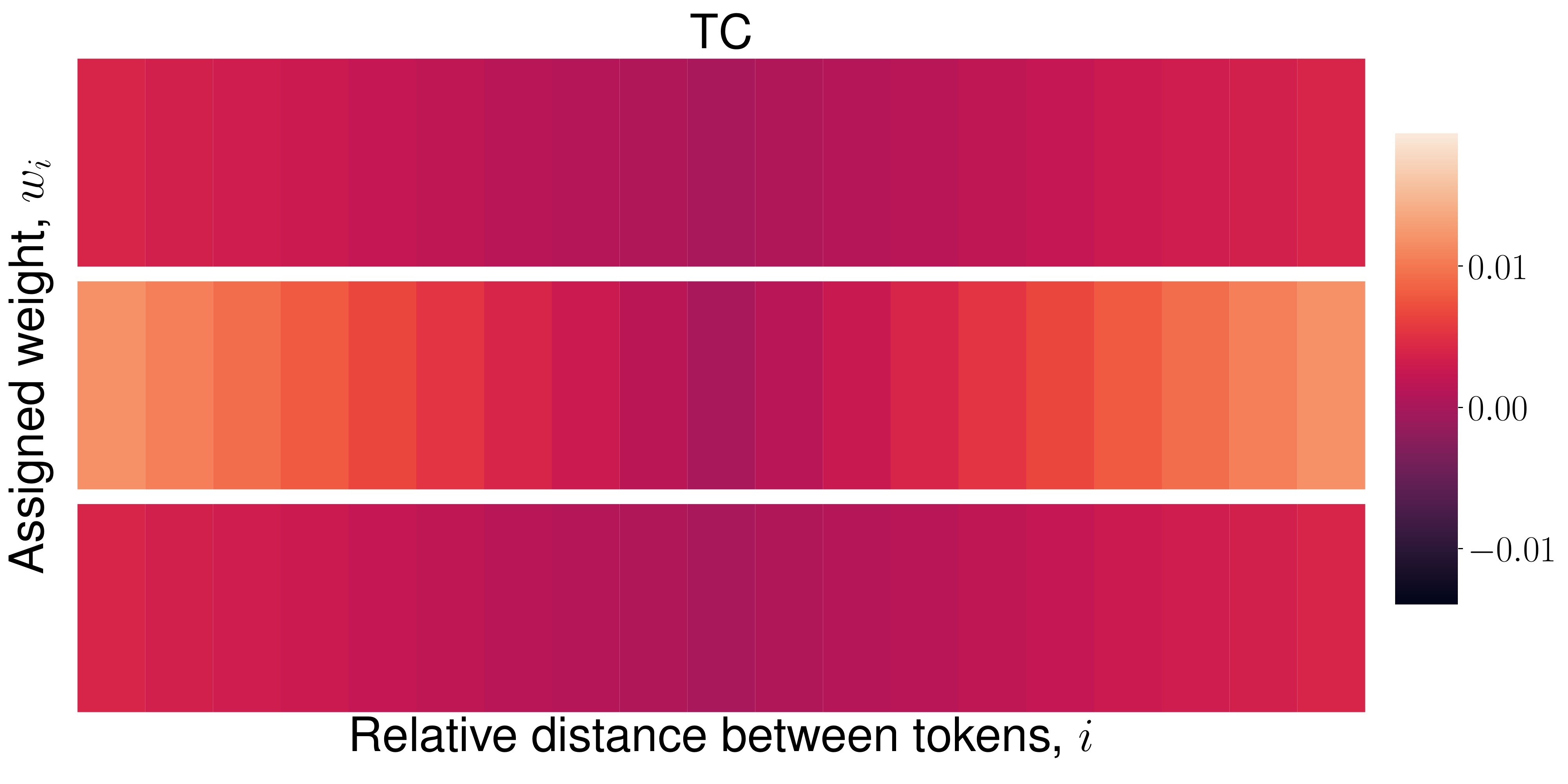}
  \caption{Weights $w_i$, learned on the TC task, assigned to the pairwise distances between tokens $i$ in FastRPB 1D.}
    \label{fig:tc}
\end{subfigure}
\end{figure}

\section{Proposition Proofs}
\subsection{Circulant Matrices}
\label{circulant}

To design a more efficient positional encoding method, we leveraged circulant matrices, which are a subclass of matrices with special properties due to their relation to the Fast Fourier Transform (FFT) and circular convolution~\cite{bamieh2020discovering}. Here, we will only focus on the property that allows calculating a matrix-vector product quickly and efficiently in terms of speed and memory. Given a vector $\vc = (c_0, c_1, ..., c_{n-1})$, we will define the associated $n \times n$ circulant matrix $\mC = \text{circ}(\vc)$
in which the first column is exactly $\vc$, and each subsequent column is obtained by a circular shift of the previous column: 
\begin{align}
    \mC = \begin{pmatrix}
    c_0     & c_{n-1} & c_{n-2} & \cdots & c_{1}  \\
	c_{1}   & c_0     & c_{n-1} &        & c_{2}  \\
	c_{2}   & c_{1}   & c_0     &        & c_{3}  \\
	\vdots  &         & \ddots  & \ddots & \vdots \\
	c_{n-1} & c_{n-2} & c_{n-3} & \cdots 
    \end{pmatrix}
\end{align}
For every vector $\vx$ of size $n$, the matrix-vector product $\mC\vx$ requires only $\mathcal{O}(n \log{n})$ computation~\cite{rosowski2021fast}. In addition, to compute the above product, it is not necessary to store the whole matrix $\mC$ in memory, as it is enough to only keep $\mathcal{O}(n)$ parameters of vector $\vc$. We will prove that FastRPB and FastRPB 2D can be expressed through circular matrices, and hence relative positional information can be embedded efficiently in the self-attention mechanism.

\subsection{Proposition 1}
\label{proof1}

In~\ref{toeplitz_matrix} we introduced a Toeplitz matrix $\mW_{1d}$ of shape $N \times N$:

\begin{align}
    \mW_{1d} = \begin{pmatrix} w_{0} & w_{1} & w_2 & \cdots & w_{N-1} \\ w_{-1} & w_{0} & w_1 & \cdots & w_{N-2} \\ w_{-2} & w_{-1} & w_0 & & w_{N-3} \\ \vdots & & \ddots & \ddots & \vdots \\ w_{-N+1} & w_{-N+2} & w_{-N+3} & \cdots & w_{0} \end{pmatrix}
\end{align}

Our goal is to efficiently multiply $\mW_{1d}$ by an arbitrary matrix $\mV$ of shape $N \times D$. As was stated in section~\nameref{circulant}, a special class of matrices, namely circulant matrices, can be multiplied by a vector efficiently in $\mathcal{O}(N \log{N})$ operations, requiring $\mathcal{O}(N)$ memory. We will extend matrix $\mW_{1d}$ with additional rows and columns and thus obtain a circulant matrix $\mW_{1d}^{\text{ext}}$. Then, we will introduce $\mV^{\text{ext}}$, a modified version of matrix $\mV$, which $\mW_{1d}^{\text{ext}}$ will be multiplied by. Finally, we will select a slice from the product $\mW_{1d}^{\text{ext}}\cdot\mV^{\text{ext}}$, which will be exactly $\mW_{1d}\cdot\mV$. 

The first step is to define $\mW_{1d}^{\text{ext}}$:
\begin{align}
    \mW_{1d}^{\text{ext}} = \begin{pmatrix}
    w_{-N+1} & w_{-N+2} & \cdots & w_{0} & w_{1} & w_2 & \cdots & w_{N-1}  \\
	w_{N-1} & w_{-N+1} & \cdots & w_{-1} & w_{0} & w_1 & \cdots & w_{N-2}  \\
	w_{N-2} & w_{N-1} & \cdots & w_{-2} & w_{-1} & w_{0} &     & w_{N-3}  \\
	\vdots &         & \ddots &      &      &  \ddots & \ddots & \vdots \\
	w_{1} & w_{2} & \cdots  & w_{-N+1}  & w_{-N+2} & \cdots & w_{-1} & w_0 \\
	\vdots &         & \ddots &      & \ddots &  \ddots & \ddots & \vdots \\
	w_{-N+2} & w_{-N+3} & \cdots  & \cdots  & \cdots & \cdots & w_{N-1} & w_{-N+1} \\
    \end{pmatrix}
\end{align}

As can be seen above, the constructed matrix $\mW_{1d}^{\text{ext}}$ is indeed circulant. In addition, the right upper corner of $\mW_{1d}^{\text{ext}}$ is essentially $\mW_{1d}$. Hence, $\mW_{1d}$ can be expressed as a slice of $\mW_{1d}^{\text{ext}}$ following the numpy notation: $\mW_{1d} = \big(\mW_{1d}^{\text{ext}}\big)_{N: \ , \ :N+1}$. 

Now we want to calculate the matrix product $\mW_{1d} \cdot \mV$ using a matrix $\mW_{1d}^{\text{ext}}$ of size $(2N-1) \times (2N-1)$. Due to this fact, we will need to multiply $\mW_{1d}^{\text{ext}}$ with an appropriate matrix $\mV^{\text{ext}}$ of size $2N-1 \times D$. As seen before, $\mW_{1d}$ is a slice of $\mW_{1d}^{\text{ext}}$, which is why we only need the first $N$ rows of the resulting product $\mW_{1d}^{\text{ext}}\cdot\mV^{\text{ext}}$. In other words, we need to find such a matrix $\mV^{\text{ext}}$ that $\mW_{1d} \cdot \mV = \big(\mW_{1d}^{\text{ext}} \cdot \mV^{\text{ext}}\big)_{:N+1 \ , \ :}$. To achieve this, we can pad $\mV$ with $N-1$ additional rows filled with zeros:
\begin{align}
    \mV^{\text{ext}} = \begin{pmatrix}
    0 & 0 & \cdots & 0 \\
    0 & 0 & \cdots & 0  \\
    \vdots &  \vdots &  \ddots &  \vdots  \\
    0 & 0 & \cdots & 0  \\
	v_{0, 0} & v_{0, 1} & \cdots & v_{0, D-1}\\
	v_{1, 0} & v_{1, 1} & \cdots & v_{1, D-1}\\
	\vdots &  \vdots &  \ddots &  \vdots  \\
	v_{N-1, 0} & v_{N-1, 1} & \cdots & v_{N-1, D-1}\\
    \end{pmatrix}
\end{align}
To complete the proof, let's explicitly show that $\mW_{1d} \cdot \mV = \big(\mW_{1d}^{\text{ext}} \cdot \mV^{\text{ext}}\big)_{:N+1 \ , \ :}$. Let's assume $D=1$, and the generalization for bigger dimensions can be done using similar operations:
\begin{align}
    \big(\mW_{1d}^{\text{ext}}\cdot\mV^{\text{ext}}\big)_{:N+1 \ , \ :} = \begin{pmatrix}
    w_{-N+1} & \cdots & w_{0} & w_{1}& \cdots & w_{N-1}  \\
	w_{N-1} &  \cdots & w_{-1} & w_{0} & \cdots & w_{N-2}  \\
	w_{N-2} & \cdots & w_{-2} & w_{-1} &      & w_{N-3}  \\
	\vdots & \ddots &      &      &  \ddots & \vdots \\
	w_{1} & \cdots  & w_{-N+1}  & w_{-N+2} & \cdots  & w_0 \\
    \end{pmatrix} \begin{pmatrix}
    0 \\
    0 \\
    \vdots \\
    0 \\
	v_{0}\\
	v_{1}\\
	\vdots\\
	v_{N-1}\\
    \end{pmatrix} = 
\end{align}
\begin{align}
    \begin{pmatrix}
    w_{0} & w_{1}& \cdots & w_{N-1}  \\
	w_{-1} & w_{0} & \cdots & w_{N-2}  \\
	w_{-2} & w_{-1} &      & w_{N-3}  \\
	\vdots & \ddots & \ddots & \vdots \\
	w_{-N+1}  & w_{-N+2} & \cdots  & w_0 \\
    \end{pmatrix} \begin{pmatrix}
	v_{0}\\
	v_{1}\\
	\vdots\\
	v_{N-1}\\
    \end{pmatrix} = \mW_{1d}\cdot\mV
\end{align}
The last thing we have to do is to calculate the complexity of the matrix product of the circulant matrix $\mW_{1d}^{\text{ext}}$ with $\mV^{\text{ext}}$. Since the matrix $\mW_{1d}^{\text{ext}}$ is circulant of size $(2N-1) \times (2N-1)$, according to the section~\nameref{circulant}, it requires $O(N)$ memory and $\mathcal{O}(N\log{N})$ operations to perform a matrix-vector product with vector of size $2N-1$. In this regard, to compute a matrix product with matrix $\mV$ of size $N \times D$, we will need to perform $D$ times more operations, i.e. $\mathcal{O}(DN\log{N})$ operations.

\subsection{Proposition 2}
\label{proof2}

In the following sections, we will be considering a $3 \times 3$ image. In section~\nameref{proof2:struct}, we will study the general structure of tensors $\tX$ and $\tY$, which were introduced in section~\nameref{prop:toeplitz_2d}. In section~\nameref{proof2:flat}, we will reshape these tensors and obtain a new pair of tensors $\mX_{\text{flat}}$ and $\mY_{\text{flat}}$, which will be then efficiently multiplied by $\mV$ matrix in the final section~\nameref{proof2:efficient}.

\subsubsection{Structure of Pairwise Distance Tensors}
\label{proof2:struct}

Firstly, to gain a deeper understanding of the structure of tensors $\tX$ and $\tY$, we will explicitly write down the components of these tensors for an image of size $3 \times 3$. Consider $\mX_{1,1,:,:}$ and $\mY_{1,1,:,:}$, which contain the weights assigned to the vertical and horizontal relative distances from pixel $(1, 1)$ to all other pixels:
\begin{align}
    \mX_{1,1,:,:} = \begin{pmatrix}
        w_{-1} & w_{-1} & w_{-1} \\
        w_0 & w_0 & w_0 \\
        w_{1} & w_{1} & w_{1}
    \end{pmatrix}, \
    \mY_{1,1,:,:} = \begin{pmatrix}
        w_{-1} & w_0 & w_{1} \\
        w_{-1} & w_0 & w_{1} \\
        w_{-1} & w_{0} & w_{1}
    \end{pmatrix}
\end{align}
$\mX_{n, m, :, :}$ and $\mY_{n, m, :, :}$ have the symmetry property, through which it can be proven that:
\begin{restatable}[]{proposition}{Symmetry}
    \label{prop:symmetry}
    $\mX_{n,i,:,:} = \mX_{n,j,:,:}$ and $\mY_{i,n,:,:} = \mY_{j,n,:,:}$ for every $n, i, j \in \{0, ..., N-1\}$, and this property holds for images of any size.
\end{restatable} 
Taking advantage of the proposition~\ref{prop:symmetry}, we can write out the explicit form of tensors $\tX$ and $\tY$ in case of $3 \times 3$ images. We will introduce the following notation:
\begin{align}
    \mA = \begin{pmatrix}
        w_{-2} & w_{-2} & w_{-2} \\
        w_{-1} & w_{-1} & w_{-1} \\
        w_0 & w_0 & w_0 \\
    \end{pmatrix}, \
    \mB = \begin{pmatrix}
        w_{-1} & w_{-1} & w_{-1} \\
        w_0 & w_0 & w_0 \\
        w_{1} & w_{1} & w_{1} \\
    \end{pmatrix}, \
    \mC = \begin{pmatrix}
        w_0 & w_0 & w_0 \\
        w_{1} & w_{1} & w_{1} \\
        w_{2} & w_{2} & w_{2} \\
    \end{pmatrix} 
\end{align}
It can be seen that different slices of the tensor $\tX$ can be expressed using matrices $\mA, \mB, \mC$:
\begin{align}
    \mX_{0, 0, :, :} = \mX_{0, 1, :, :} = \mX_{0, 2, :, :} = \mC \\
    \mX_{1, 0, :, :} = \mX_{1, 1, :, :} = \mX_{1, 2, :, :} = \mB \\
    \mX_{2, 0, :, :} = \mX_{2, 1, :, :} = \mX_{2, 2, :, :} = \mA
\end{align}
Moreover, are matrices $\mA, \mB, \mC$ also applicable for tensor $\tY$:
\begin{align}
    \mY_{0, 0, :, :} = \mY_{0, 1, :, :} = \mY_{0, 2, :, :} = \mC^T \\
    \mY_{1, 0, :, :} = \mY_{1, 1, :, :} = \mY_{1, 2, :, :} = \mB^T \\
    \mY_{2, 0, :, :} = \mY_{2, 1, :, :} = \mY_{2, 2, :, :} = \mA^T
\end{align}

\subsubsection{Flattening of Tensors}
\label{proof2:flat}

In the transformer architecture, before processing the image of size $N \times N$, it is usually flattened into a one-dimensional vector of size $N^2$. To define the flattening operation, consider an arbitrary matrix $\mM$ of shape $3 \times 3$. Its flattened version $\vm_{\text{flat}}$ will have the following structure:
\begin{align}
    \mM = \begin{pmatrix} 
    m_{0,0} & m_{0,1} & m_{0,2} \\
    m_{1,0} & m_{1,1} & m_{1,2} \\
    m_{2,0} & m_{2,1} & m_{2,2} \\
    \end{pmatrix} \ \xrightarrow{\text{flattening}} \ \vm_{\text{flat}} = \begin{pmatrix} 
    m_{0,0} \\ m_{0,1} \\ m_{0,2} \\ m_{1,0} \\ m_{1,1}  \\ m_{1,2} \\ m_{2,0} \\ m_{2,1} \\ m_{2,2} \\
    \end{pmatrix}
\end{align}
Due to the flattening of images in the transformer, the matrix $\mV$ in will have a shape of $N^2 \times D$, and hence it is essential to reshape tensors $\tX$ and $\tY$ from size $(N \times N) \times (N \times N)$ to size $N^2 \times N^2$. 
We will denote reshaped versions of tensors $\tX$ and $\tY$ as $\mX_{\text{flat}}$ and $\mY_{\text{flat}}$ respectively. Reshaping of the above tensors can be decomposed into two consecutive flattening operations, first applied to the last two dims of tensors $\tX$ and $\tY$, and then to the first two. Flattening of the last dimensions is equivalent to flattening of each matrix $\mA, \mB, \mC$. After this operation, we will obtain the following three vectors for $3 \times 3$ images:
\begin{align}
    \va_{\text{flat}}= \begin{pmatrix} 
    w_{0} \\ w_{0}  \\ w_{0}  \\ w_{1} \\ w_{1}  \\ w_{1} \\ w_{2}  \\ w_{2} \\ w_{2} \\
    \end{pmatrix}, \ \vb_{\text{flat}} = \begin{pmatrix} 
    w_{-1}  \\ w_{-1} \\ w_{-1} \\
    w_{0} \\ w_{0}  \\ w_{0}  \\ w_{1} \\ w_{1}  \\ w_{1} \\
    \end{pmatrix}, \ \vc_{\text{flat}}= \begin{pmatrix} w_{-2}  \\ w_{-2} \\ w_{-2} \\
    w_{-1}  \\ w_{-1} \\ w_{-1} \\
    w_{0} \\ w_{0}  \\ w_{0}
    \end{pmatrix}
\end{align}
After the next flattening operation, we will get the following:
\begin{align}
\mX_{\text{flat}} = \tX.\text{reshape}(N^2, N^2) = \begin{pmatrix}  
    w_0 & w_0 & w_0 & w_1 & w_1 & w_1 & w_2 & w_2 & w_2\\  
    w_0 & w_0 & w_0 & w_1 & w_1 & w_1 & w_2 & w_2 & w_2\\  
    w_0 & w_0 & w_0 & w_1 & w_1 & w_1 & w_2 & w_2 & w_2\\   
    w_{-1} & w_{-1} & w_{-1} & w_0 & w_0 & w_0 & w_1 & w_1 & w_1 \\
    w_{-1} & w_{-1} & w_{-1} & w_0 & w_0 & w_0 & w_1 & w_1 & w_1 \\
    w_{-1} & w_{-1} & w_{-1} & w_0 & w_0 & w_0 & w_1 & w_1 & w_1 \\
    w_{-2} & w_{-2} & w_{-2}  & w_{-1} & w_{-1} & w_{-1} & w_0 & w_0 & w_0 \\ 
    w_{-2} & w_{-2} & w_{-2}  & w_{-1} & w_{-1} & w_{-1} & w_0 & w_0 & w_0 \\
    w_{-2} & w_{-2} & w_{-2}  & w_{-1} & w_{-1} & w_{-1} & w_0 & w_0 & w_0 \\
    \end{pmatrix}
\label{x_flat}
\end{align}

\begin{align}
\mY_{\text{flat}} = \tY.\text{reshape}(N^2, N^2) =
\begin{pmatrix}       
    w_0 & w_1 & w_2 & w_0 & w_1 & w_2 & w_0 & w_1 & w_2\\ 
    w_{-1} & w_0 & w_1 & w_{-1} & w_0 & w_1 & w_{-1} & w_0 & w_1\\ 
    w_{-2} & w_{-1} & w_0 & w_{-2} & w_{-1} & w_0 & w_{-2} & w_{-1} & w_0\\ 
    w_0 & w_1 & w_2 & w_0 & w_1 & w_2 & w_0 & w_1 & w_2\\ 
    w_{-1} & w_0 & w_1 & w_{-1} & w_0 & w_1 & w_{-1} & w_0 & w_1\\ 
    w_{-2} & w_{-1} & w_0 & w_{-2} & w_{-1} & w_0 & w_{-2} & w_{-1} & w_0\\ 
    w_0 & w_1 & w_2 & w_0 & w_1 & w_2 & w_0 & w_1 & w_2\\ 
    w_{-1} & w_0 & w_1 & w_{-1} & w_0 & w_1 & w_{-1} & w_0 & w_1\\ 
    w_{-2} & w_{-1} & w_0 & w_{-2} & w_{-1} & w_0 & w_{-2} & w_{-1} & w_0\\ 
    \end{pmatrix}
\label{y_flat}
\end{align}
Note that the rows of $\mX_{\text{flat}}$ and $\mY_{\text{flat}}$ are $\va_{\text{flat}}, \vb_{\text{flat}}$ and $\vc_{\text{flat}}$.

Now that we have the flattened representations of $\mX_{\text{flat}}$ and $\mY_{\text{flat}}$, the last step we have to take is to compute the matrix product $\mX_{\text{flat}}\cdot\mV$ and $\mY_{\text{flat}}\cdot\mV$.

\subsubsection{Efficient Matrix Product}
\label{proof2:efficient}

$\textbf{Calculation of } \mY_{\text{flat}}\cdot\mV$

In case of $3 \times 3$ images, it is easy to see that the matrix $\mY_{\text{flat}}$, presented in formula~\ref{y_flat} consists of 9 identical blocks. We will denote these blocks as $\mR$:
\begin{align}
    \mY_{\text{flat}} = \begin{pmatrix} 
        \mR & \mR & \mR \\ 
        \mR & \mR & \mR \\ 
        \mR & \mR & \mR 
    \end{pmatrix}, \text{ where } \
    \mR = \begin{pmatrix} 
        w_0 & w_1 & w_2 \\  
        w_{-1} & w_0 & w_1 \\  
        w_{-2} & w_{-1} & w_0 
    \end{pmatrix} 
\end{align}
Note that the matrix $\mR$ is just a block of $\mY$, not its element. Another very important fact is that the matrix $\mR$ is a Toeplitz. Therefore, according to~\nameref{proof1}, it can be efficiently multiplied by a vector. Our goal is to efficiently multiply $\mY_{\text{flat}}$ with a matrix $\mV$ of shape $N^2 \times D$. Since $\mV$ is basically vectors of size $N^2$ stacked $D$ times, it will be enough to consider the matrix-vector product of $\mY_{\text{flat}}$ with a vector $\vv$ of shape $N^2$. The product can be expressed as:
\begin{align}
    \mY_{\text{flat}}\cdot \mV = \begin{pmatrix} 
        \mR & \mR & \mR \\ 
        \mR & \mR & \mR \\ 
        \mR & \mR & \mR 
    \end{pmatrix} \begin{pmatrix}
        v_0 \\ v_1 \\ v_2 \\ v_3 \\ v_4 \\ v_5 \\ v_6 \\ v_7 \\ v_8
    \end{pmatrix} = \begin{pmatrix} 
        \mR \cdot \begin{pmatrix}
        v_0 \\ v_1 \\ v_2
        \end{pmatrix} + \mR \cdot \begin{pmatrix}
        v_3 \\ v_4 \\ v_5
        \end{pmatrix}+ \mR \cdot \begin{pmatrix}
        v_6 \\ v_7 \\ v_8
        \end{pmatrix} \\ 
        \vdots \\
        \vdots \\
    \end{pmatrix} =
\end{align}
\begin{align}
\label{yv_prod}
     \begin{pmatrix} 
        \mR \cdot \begin{pmatrix}
        v_0+v_3+v_6 \\ v_1+v_4+v_7 \\ v_2+v_5+v_8
        \end{pmatrix}  \\ 
        \vdots \\
        \vdots \\
    \end{pmatrix}
\end{align}
Let's introduce an additional notation: we will denote as $\mV_m$ a tensor of size $N \times N \times D$, which was obtained through reshaping a matrix $\mV$ of size $N^2 \times D$. $\mV_m$ will have the following structure:
\begin{align}
\label{vm}
    \mV_m = \begin{pmatrix} v_0 & v_1 & v_2 \\ v_3 & v_4 & v_5 \\ v_6 & v_7 & v_8\end{pmatrix} = \mV.\text{reshape(N, N, D)}
\end{align}
Then:
\begin{align}
    \mV_m^T.\text{sum}(1) = \begin{pmatrix} v_0 + v_3 + v_6 \\ v_1 + v_4 + v_7 \\ v_2 + v_5 + v_8\end{pmatrix}
\end{align}
Finally using notation~\ref{vm} and result of product~\ref{yv_prod}, we can conclude that:
\begin{align}
\mY_{\text{flat}}\cdot\mV = \begin{pmatrix} \mR \cdot \Big(\mV_m^T.\text{sum}(1)\Big) \\  \vdots \\ \vdots \end{pmatrix}
\end{align}
Note that the summation is by $\text{dim}=1$, not by $\text{dim}=-1$. That is because in general cases, when $D$ is not 1, the latter dimension refers to the hidden size.

$\mR$ is a Topleitz matrix we worked with in the previous paragraph~\ref{proof1}. That is, it is the product of the matrix $\mR$ with $\mV_m^T.\text{sum}(1)$ can be efficiently computed using the properties of Toeplitz matrix. Finally, we can conclude that a $N \times N$ image computation of product $\mY_{\text{flat}}\cdot\mV$ requires $\mathcal{O}(DN\log{N})$ computations and $\mathcal{O}(N)$ memory to store a vector, generating matrix $\mR$.

$\textbf{Calculation of } \mX_{\text{flat}}\cdot\mV$

Here, we will rearrange the columns of matrix $\mX_{\text{flat}}$ to construct a matrix $\mX_{\text{flat}}'$, and construct matrix $\mU$ accordingly by rearranging the rows of matrix $\mV$ in such a way that the equality holds $\mX_{\text{flat}}\cdot\mV=\mX_{\text{flat}}'\cdot\mU$. We still will work with the $D=1$ case since, as was noted previously, it can be easily generalized for higher dimensions.
    
\begin{align}
    \mX_{\text{flat}}\cdot\mV = \begin{pmatrix}  
    w_0 & w_0 & w_0 & w_1 & w_1 & w_1 & w_2 & w_2 & w_2\\  
    w_0 & w_0 & w_0 & w_1 & w_1 & w_1 & w_2 & w_2 & w_2\\  
    w_0 & w_0 & w_0 & w_1 & w_1 & w_1 & w_2 & w_2 & w_2\\   
    w_{-1} & w_{-1} & w_{-1} & w_0 & w_0 & w_0 & w_1 & w_1 & w_1 \\
    w_{-1} & w_{-1} & w_{-1} & w_0 & w_0 & w_0 & w_1 & w_1 & w_1 \\
    w_{-1} & w_{-1} & w_{-1} & w_0 & w_0 & w_0 & w_1 & w_1 & w_1 \\
    w_{-2} & w_{-2} & w_{-2}  & w_{-1} & w_{-1} & w_{-1} & w_0 & w_0 & w_0 \\ 
    w_{-2} & w_{-2} & w_{-2}  & w_{-1} & w_{-1} & w_{-1} & w_0 & w_0 & w_0 \\
    w_{-2} & w_{-2} & w_{-2}  & w_{-1} & w_{-1} & w_{-1} & w_0 & w_0 & w_0 \\
    \end{pmatrix}\cdot \begin{pmatrix}v_0\\v_1\\ v_2\\ v_3\\ v_4\\ v_5\\ v_6\\ v_7\\ v_8\end{pmatrix} =
\end{align}
\begin{align}
    \begin{pmatrix}  
    w_0 & w_1 & w_2 & w_0 & w_1 & w_2 & w_0 & w_1 & w_2\\  
    w_0 & w_1 & w_2 & w_0 & w_1 & w_2 & w_0 & w_1 & w_2\\  
    w_0 & w_1 & w_2 & w_0 & w_1 & w_2 & w_0 & w_1 & w_2\\   
    w_{-1} & w_{0} & w_{1} & w_{-1} & w_0 & w_1 & w_{-1} & w_0 & w_1 \\
    w_{-1} & w_{0} & w_{1} & w_{-1} & w_0 & w_1 & w_{-1} & w_0 & w_1 \\
    w_{-1} & w_{0} & w_{1} & w_{-1} & w_0 & w_1 & w_{-1} & w_0 & w_1 \\
    w_{-2} & w_{-1} & w_{0}  & w_{-2} & w_{-1} & w_{0} & w_{-2} & w_{-1} & w_{0} \\ 
    w_{-2} & w_{-1} & w_{0}  & w_{-2} & w_{-1} & w_{0} & w_{-2} & w_{-1} & w_{0} \\ 
    w_{-2} & w_{-1} & w_{0}  & w_{-2} & w_{-1} & w_{0} & w_{-2} & w_{-1} & w_{0} \\ 
    \end{pmatrix} \cdot \begin{pmatrix}v_0 \\v_3\\ v_6\\ v_1\\ v_4\\ v_7\\ v_2\\ v_5\\ v_8\end{pmatrix} = \mX_{\text{flat}}'\cdot\mU = \begin{pmatrix}\alpha \\ \alpha\\ \alpha\\ \beta \\ \beta \\ \beta\\ \gamma\\ \gamma\\ \gamma \end{pmatrix}
\end{align}
Now let's use the already introduced matrix $\mV_m$:
\begin{align}
\mR \cdot\big(\mV_m.\text{sum}(1)\big) = \begin{pmatrix} 
        w_0 & w_1 & w_2 \\  
        w_{-1} & w_0 & w_1 \\  
        w_{-2} & w_{-1} & w_0 
    \end{pmatrix} \begin{pmatrix}
        v_0 + v_1 + v_2 \\ v_3 + v_4 + v_5 \\ v_6 + v_7 + v_8
    \end{pmatrix} =
\end{align}
\begin{align}
     \mR \cdot \begin{pmatrix}
        v_0 \\ v_3 \\ v_6
    \end{pmatrix} + \mR \cdot \begin{pmatrix}
        v_1 \\ v_4 \\ v_7
    \end{pmatrix} + \mR \cdot \begin{pmatrix}
        v_2 \\ v_5 \\ v_8
    \end{pmatrix} = \begin{pmatrix} \alpha \\ \beta \\ \gamma \end{pmatrix}
\end{align}
And finally:
\begin{align}
     \mX_{\text{flat}}'\cdot\mU = \begin{pmatrix}\alpha \\ \alpha\\ \alpha\\ \beta \\ \beta \\ \beta\\ \gamma\\ \gamma\\ \gamma \end{pmatrix} = \begin{pmatrix}\alpha & \alpha & \alpha\\ \beta & \beta & \beta\\ \gamma & \gamma & \gamma \end{pmatrix}.\text{reshape}(N^2) =
\end{align}
\begin{align}
    \begin{pmatrix}
    \mR \cdot \big(\mV_m.\text{sum}(1)\big) & \cdots & \cdots
    \end{pmatrix}.\text{reshape}(N^2)
\end{align}

\subsubsection{Conclusion}
As we proved in the previous section, products $\mX_{\text{flat}}\cdot\mV$ and $\mY_{\text{flat}}\cdot\mV$ can be computed using just two products: $\mR\cdot\Big(\mV_m.\text{sum}(1)\Big)$ and $\mR\cdot\Big(\mV_m^T.\text{sum}(1)\Big)$ respectively. Moreover, matrix $\mR$ utilized in the above products is a Toeplitz, and in this regard this products can be computed efficiently according to~\nameref{proof1}. To summarize, we prove that FastRBP 2D will require be $\mathcal{O}(DN\log{N})$ computations and $\mathcal{O}(N)$ memory.

\end{document}